\documentclass[11pt]{article}

\usepackage[final]{acl}
\usepackage{graphicx}
\usepackage{subcaption} 
\usepackage{times}
\usepackage{latexsym}
\usepackage{graphicx}
\usepackage{amsmath}
\usepackage{amssymb}
\usepackage{booktabs}
\usepackage{adjustbox}
\usepackage[table]{xcolor}
\usepackage{algorithm}
\usepackage{tikz}
\usetikzlibrary{arrows.meta}
\usepackage{algorithmic}
\usepackage{xcolor} 
\usepackage{times}
\usepackage{latexsym}
\usepackage{amsmath}
\usepackage{multirow}
\usepackage{svg}
\usepackage{xcolor}
\usepackage{colortbl}
\usepackage{makecell} 
\usepackage{caption}
\usepackage{subcaption}
\usepackage{booktabs}
\usepackage{inconsolata}
\usepackage{xcolor}
\usepackage{enumerate}
\usepackage{graphicx}
\usepackage{amssymb}
\usepackage{bbm}
\usepackage{times}
\usepackage{latexsym}
\usepackage{graphicx}
\usepackage[T1]{fontenc}

\usepackage[utf8]{inputenc}

\usepackage{microtype}

\usepackage{inconsolata}

\usepackage{graphicx}

\usepackage{cancel}

\newcommand\todo[1]{\textcolor{red}{#1}}
\newcommand{\ourmethod}{MiMIC}
\newcommand{\imgC}{\mathcal I^{C}}
\newcommand{\imgV}{\mathcal I^{V}}
\newcommand{\imgVC}{\mathcal I^{VC}}
\newcommand{\textDoc}{\mathcal T}
\newcommand{\corpusIvc}{\mathcal{D}_{I^{VC}}}
\newcommand{\corpusIv}{\mathcal{D}_{I^{V}}}

\newcommand{\UniVLDR}{UniVL-DR}
\newcommand{\Marvel}{Marvel}

\newcommand{\corpusall}{\mathcal{D}_{\{T,I^V,I^{VC}\}}}

\title{\ourmethod{}: Mitigating Visual Modality Collapse in \\ Universal Multimodal Retrieval While Avoiding Semantic Misalignment}

\author{
 \textbf{Li Juan\textsuperscript{1,2}},
 \textbf{Chuanghao Ding\textsuperscript{1,2}},
 \textbf{Xujie Zhang\textsuperscript{1,2}},
 \textbf{Cam-Tu Nguyen\thanks{Corresponding Author: ncamtu@nju.edu.cn} \textsuperscript{1,2}}       
\\
 \textsuperscript{1}State Key Laboratory for Novel Software Technology, Nanjing University,
 \\
 \textsuperscript{2}School of Artificial Intelligence, Nanjing University
\\
\texttt{juanli@smail.nju.edu.cn}
 }

\begin{document}

\maketitle

\begin{abstract}
Universal Multimodal Retrieval (UMR) aims to map different modalities (e.g., visual and textual) into a shared embedding space for multi-modal retrieval. Existing UMR methods can be broadly divided into two categories: \textit{early-fusion approaches}, such as \Marvel{}, which projects visual features into the language model (LM) space for integrating with text modality, and \textit{late-fusion approaches}, such as \UniVLDR{}, which encode visual and textual inputs using separate encoders and obtain fused embeddings through addition.
Our pilot study reveals that \Marvel{} exhibits \textit{visual modality collapse}, which is characterized by the model’s tendency to disregard visual features while depending excessively on textual cues. In contrast, although \UniVLDR{} is less affected by this issue, it is more susceptible to \textit{semantic misalignment}, where semantically related content is positioned far apart in the embedding space. To address these challenges, we propose \textbf{\ourmethod{}}, which introduces two key innovations: (1) a fusion-in-decoder architecture for effective multimodal integration, and (2) robust training through single-modality mix-in and random caption dropout. Experiments on the WebQA+ and EVQA+ datasets—where image in documents or queries might lack captions—indicate that \ourmethod{} consistently outperforms both early- and late-fusion baselines.
\end{abstract}

\section{Introduction}
Universal Multimodal Retrieval (UMR) aims to map diverse modalities—such as images and text—into a unified, shared embedding space to facilitate seamless multi-modal search and retrieval \cite{zhou-etal-2024-marvel, liuuniversal, zhang2025gmeimprovinguniversalmultimodal}. The necessity of UMR arises from the increasingly heterogeneous nature of digital information; users frequently seek answers that require data from diverse modal sources, where a query in one modality must effectively retrieve relevant content in another. 
In addition to cross-modal retrieval, both queries and documents in UMR can be multimodal; for example, a query may consist of an image paired with a textual question.

\begin{figure}[t]
  \centering
  \includegraphics[width=\linewidth]{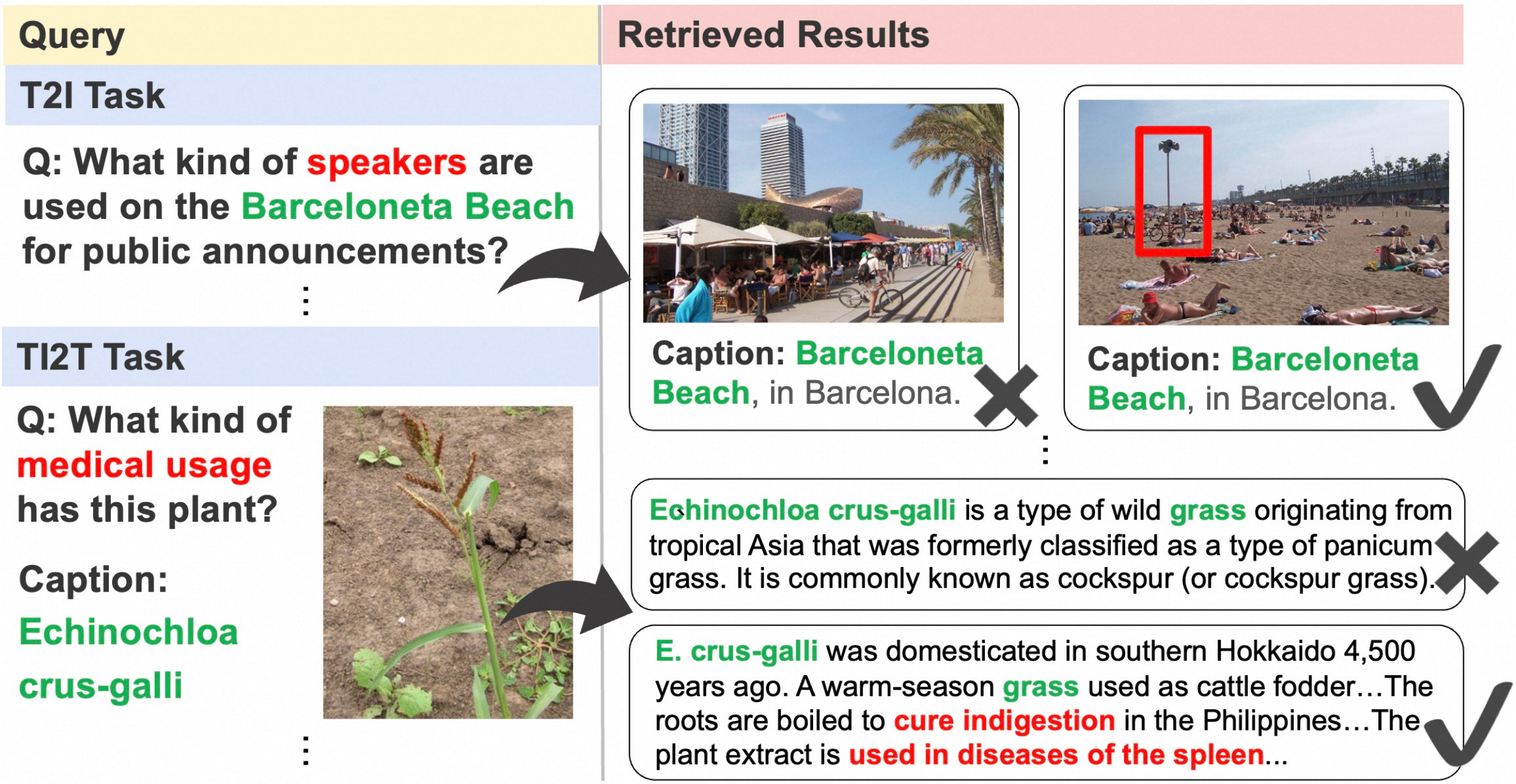}
  \caption{First row demonstrates the visual modality collapse in image document. Second row shows the semantic misalignment, that limits the model to effectively combine and understand information. }
  \label{fig:case}
\end{figure}

Existing UMR methods can be broadly categorized into two types based on their fusion strategy. Early-fusion approaches project visual features into the language model (LM) space to facilitate multi-modal interactions via self-attention mechanisms. Representative methods in this category include \Marvel{} \cite{zhou-etal-2024-marvel}, VISTA \cite{zhou-etal-2024-vista}, and GME \cite{zhang2025gmeimprovinguniversalmultimodal}. Conversely, late-fusion approaches—such as \UniVLDR{} \cite{liuuniversal}—encode visual and textual inputs using separate encoders and derive fused representations through addition.

While early-fusion methods generally outperform late-fusion alternatives, our pilot study with \Marvel{} reveals that this approach exhibits visual modality collapse. This phenomenon is characterized by the model’s tendency to disregard visual features in favor of an excessive reliance on textual cues. This observation is consistent with recent studies highlighting the ``text dominance'' issue in Visual Language Models (VLM) and Multimodal LLMs (MLLMs) \cite{wu2025language,zheng2025mllms}. In contrast, while late-fusion models like \UniVLDR{} are less prone to modality collapse, they are more susceptible to semantic misalignment, where the position of a document in semantic space is heavily influenced by its modality. Figure \ref{fig:case} illustrates these challenges: the first row depicts modality collapse, where ignored visual information regarding the \textit{``speaker''} leads to incorrect retrieval, while the second row demonstrates the model’s limited capacity to fully understand the query due to semantic misalignment.

To address these challenges, we propose \ourmethod{}, which introduces two key innovations: (1) a fusion-in-decoder architecture for effective multimodal integration, and (2) a robust training strategy incorporating single-modality mix-in and random caption dropout. Our method employs separate encoders for different modalities and utilizes cross-attention of the LM decoder to selectively aggregate relevant information from multiple modalities into the fused embedding (fusion-in-decoder). During training, we explicitly maintain and ``mix in'' single-modality representations with the fused embeddings. This design is motivated by the insight that different modalities generalize at different rates \cite{chaudhuri2025closer, Wang_2020_CVPR}. By preserving these individual modality signals, we prevent the model from discarding critical modality-specific information. Furthermore, our caption dropout strategy also forces the model to optimize visual embeddings, preventing over-reliance on textual features. To the best of our knowledge, this is the first work to investigate the modality collapse issue in UMR.

To evaluate our method, we extend the WebQA and EVQA datasets to WebQA+ and EVQA+ to include scenarios with missing modalities, requiring the model to understand information from each modality without over-relying on any one modality.

Our experimental results demonstrate that \ourmethod{} consistently outperforms early-fusion baselines, such as \Marvel{} and VISTA, as well as the late-fusion baseline, \UniVLDR{}. Furthermore, our ablation study validates the critical roles of both the fusion-in-decoder architecture and our robust training strategy in maintaining balanced performance across different retrieval settings.

Our contributions are summarized as follows:
\begin{itemize}
    \item We conduct an empirical study revealing that existing UMR methods suffer from two distinct failure modes: visual modality collapse in early-fusion models and semantic misalignment in late-fusion models.
    \item We propose the usage of Fusion-in-Decoder (FiD) for multi-modal fusion, which utilizes a language model (LM) and cross-attention to dynamically aggregate information from separate encoders. 
    \item We propose a robust strategy consisting of single-modality mix-in to preserve modality-specific signals and random caption dropout to force the model to utilize visual features, effectively mitigating text dominance.
    \item We demonstrate through extensive experiments that \ourmethod{} consistently outperforms competitive early- and late-fusion baselines, particularly in ``imperfect'' data settings.
\end{itemize} 

\section{Related Work}

\subsection{Universal Multi-modal Retrieval}

Multi-modal retrieval has attracted growing attention in recent years, with representative benchmarks such as OKVQA \cite{okvqa}, WebQA \cite{WebQA}, ViquAE \cite{lerner2022viquae}, Remuq \cite{luo-etal-2023-end}, and EVQA \cite{mensink23iccv}. Unlike conventional cross-modal retrieval, multi-modal retrieval demands effective representation of compositional information across modalities.

\begin{figure}[]
    \centering
    \begin{subfigure}{0.23\textwidth}
        \centering
        \includegraphics[width=\textwidth]{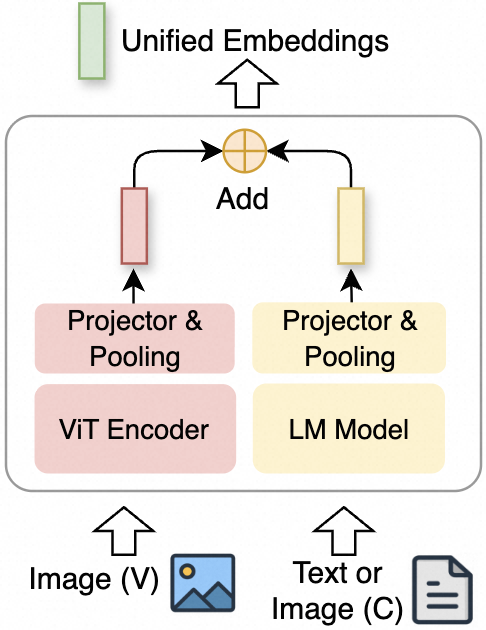}
        \caption{Late Fusion.}
        \label{fig:c}
    \end{subfigure}
    \hspace{1mm} 
    \begin{subfigure}{0.23\textwidth}
        \centering
        \includegraphics[width=\textwidth]{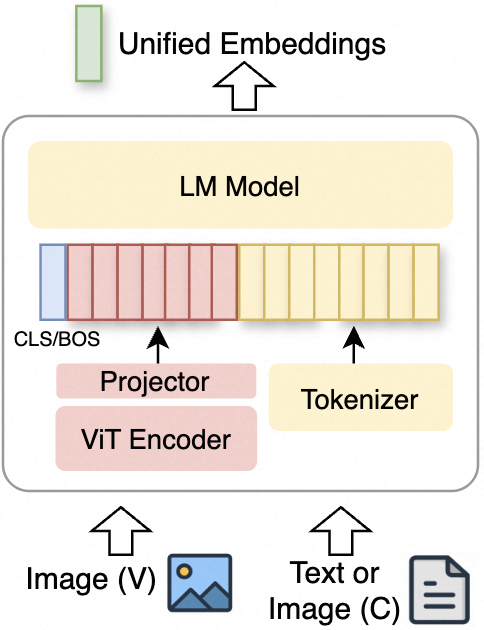}
        \caption{Early Fusion.}
        \label{fig:d}
    \end{subfigure}

    \caption{Different Fusion Strategies. (a) Late Fusion. Use separate image and text encoder (e.g., \UniVLDR{}). (b) Early Fusion. Concatenates image and text embeddings, which are then fed into the LM. (e.g., \Marvel{})}
    \label{fig:archi-others}
\end{figure}

Traditional multimodal retrieval frameworks use a pipeline, performing modality-specific retrieval with separate models before post-ranking. Text retrieval uses models like DPR \cite{dpr}, BM25 \cite{bm25} or BGE \cite{bge}, while image retrieval relies on CLIP \cite{clip}, BLIP \cite{chen-etal-2023-pre-trained} or SIGLIP \cite{zhai2023sigmoid}. This separation hinders effective cross-modal integration.

UMR maps multiple modalities into a shared semantic space for intra-, cross-, and multi-modal retrieval. Fusion is key to cross-modal integration, with models mainly using early or late fusion (Figure \ref{fig:archi-others}). Early-fusion methods, like \Marvel{} and VISTA, jointly process textual and visual tokens via self-attention for richer interactions. Late-fusion methods, such as \UniVLDR{}, rely on dual-tower VLMs (e.g., CLIP, EVA-CLIP, BLIP, SIGLIP) that encode each modality separately and fuse embeddings by addition. Unfortunately, UMR suffers from modality collapse and misalignment issues, which has not been addressed in current works.

\subsection{Modality Collapse}

Modality collapse occurs when multimodal models fail to integrate information across modalities, overfitting to a dominant modality while others lose representational capacity. These problems have been observed in multi-modal applications such as classification \cite{Wang_2020_CVPR}, question answering \cite{sim2025can} and clinical multi-modal prediction \cite{10.1145/3534678.3539388, wu2024multimodal}. In modern Vision-Language and Multimodal LLMs, this often appears as text modality dominance, where visual cues are neglected in favor of linguistic features \cite{sim2025can,chaudhuri2025closer}. This limitation is particularly concerning given that missing modalities are common in real-world.

Current studies attribute modality collapse to various factors, including multimodal polysemantic collisions \cite{chaudhuri2025closer}, dataset bias or model behavior \cite{sim2025can}, and conflicting gradients \cite{javaloy2022mitigating}. Other work highlights Transformers’ sensitivity to missing modalities and the task-dependent robustness of fusion strategies \cite{ma2022multimodal}. Despite this progress, modality collapse remains poorly understood in the context of UMR. This paper provides the first investigation addressing it from the fusion design and training strategies perspectives.

Several studies on missing-modality robustness are also relevant to our work: Ma et al. \cite{r1} investigated the robustness of multimodal transformers to missing modalities, showing that text modality often dominates in classification tasks; Liaqat et al. \cite{r2} proposed a multimodal framework (Chameleon) to address missing modalities; Malitesta et al. \cite{r3} discussed the necessity of dropping items with missing modalities in multimodal recommendation, introducing modality dropout as a common strategy for missing-modality issues. Our work differs from these studies in that while they focus on mitigating performance degradation caused by missing modalities during testing, we target modality collapse—a problem that impairs performance even when all modalities are present—for instance, when fused representations disproportionately rely on one dominant modality.

\section{Pilot Observations}
\label{sec:pilot}


\subsection{Visual Modality Collapse Issue}
\label{sec:3.1}

\begin{figure}[]
    \centering
    \begin{subfigure}{0.23\textwidth}
        \centering
        \includegraphics[width=\textwidth]{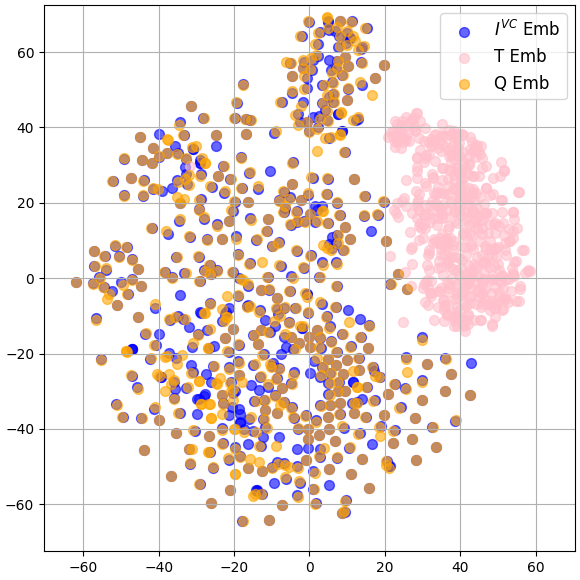}
        \caption{\Marvel{}: $Q \rightarrow \corpusIvc$}
        \label{fig:pilot-T2I-a}
    \end{subfigure}
    \hspace{1mm} 
    \begin{subfigure}{0.23\textwidth}
        \centering
        \includegraphics[width=\textwidth]{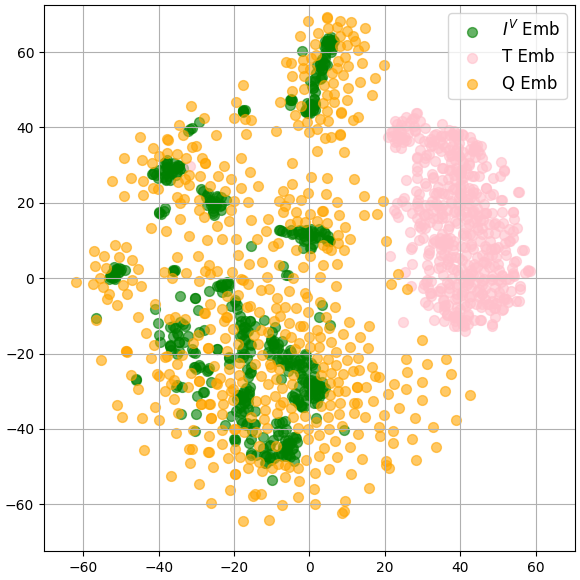}
        \caption{\Marvel{}: $Q \rightarrow\corpusIv{}$}
        \label{fig:pilot-T2I-b}
    \end{subfigure}

    \vspace{1em} 

    \begin{subfigure}{0.23\textwidth}
        \centering
        \includegraphics[width=\textwidth]{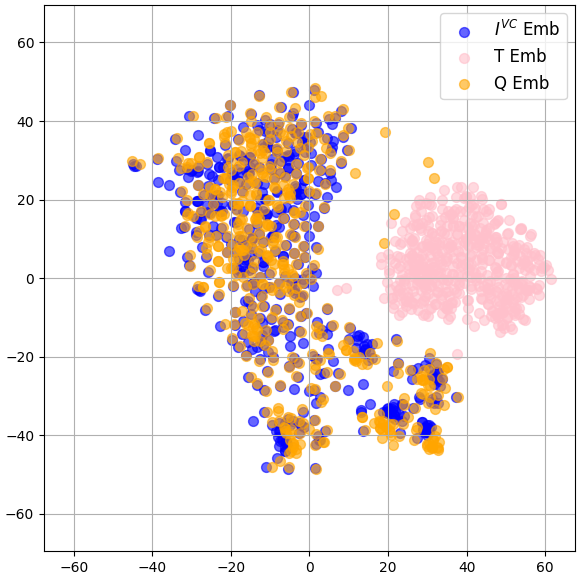}
        \caption{\UniVLDR{}: $Q \rightarrow\corpusIvc{}$}
        \label{fig:pilot-T2I-c}
    \end{subfigure}
    \hspace{1mm} 
    \begin{subfigure}{0.23\textwidth}
        \centering
        \includegraphics[width=\textwidth]{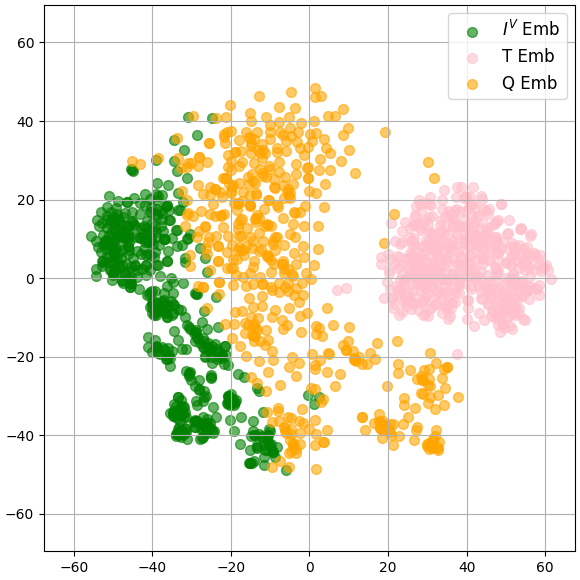}
        \caption{\UniVLDR{}: $Q \rightarrow\corpusIv{}$}
        \label{fig:pilot-T2I-d}
    \end{subfigure}

    \caption{Embeddings of T2I Queries, Image Documents ($\imgVC{}$ or $\imgV{}$) and Text Documents ($T$).}
    \label{fig:pilot-T2I}
\end{figure}

\paragraph{Experimental Setup} We trained \UniVLDR{} and \Marvel{} on the WebQA dataset. Information about the dataset is provided in Section \ref{sec:Setup}, and the implementation details for both models are given in Appendix \ref{sec:appendix1}. We then utilized the resulting models to extract query and image embeddings for the Text-to-Image (T2I) retrieval task. In this setup, the ground-truth documents are images, and the corpus exclusively consists of image documents. We evaluated two specific scenarios: retrieval in $\corpusIvc{}$, where documents $\imgVC{}$ include both visual content and captions, and $\corpusIv{}$, where documents contain only visual information. Finally, we employed t-SNE \cite{JMLR:v9:vandermaaten08a} to project the embeddings into a two-dimensional space for visualization in Figure \ref{fig:pilot-T2I}.

\paragraph{Results and Discussion} As illustrated in Figure \ref{fig:pilot-T2I}, the queries consistently cluster around their corresponding image documents across all cases, indicating effective query-document alignment for the T2I retrieval task. In the multimodal scenario where images are paired with captions (Figure \ref{fig:pilot-T2I-a}, \ref{fig:pilot-T2I-c}), the fused embeddings exhibit a broad distribution throughout the vector space. However, in the absence of captions (Figure \ref{fig:pilot-T2I-b}, \ref{fig:pilot-T2I-d}), a clear problem emerges: \Marvel{} embeddings collapse into several highly dense clusters, whereas \UniVLDR{} maintains a broad spread and more diverse representation of the visual data. 


\begin{figure}
    \centering
    \includegraphics[width=0.88\linewidth]{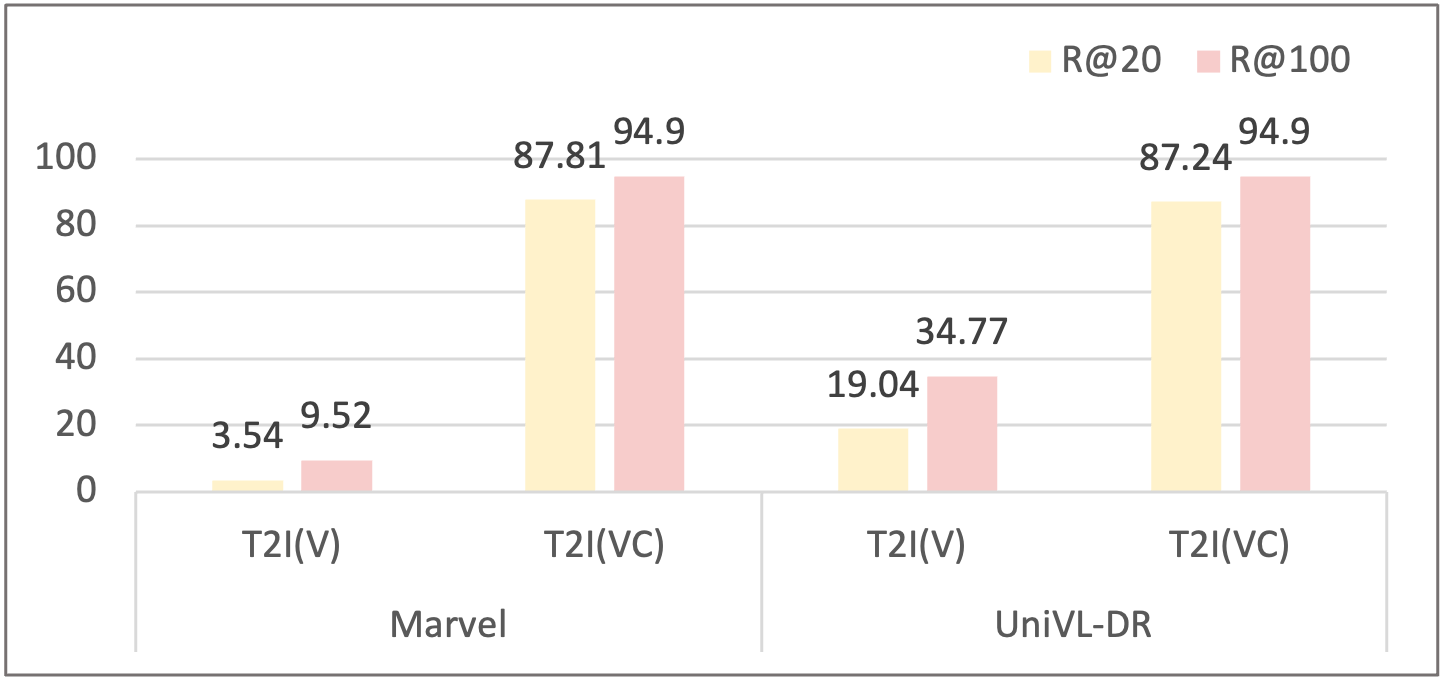}
    \caption{$T2I^V$ and $T2I^{VC}$ using Marvel and \UniVLDR{}: $T2I^{VC}$ retrieves from $\corpusIvc{}$, a corpus of images with captions; $T2I^{V}$ retrieves from $\corpusIv{}$, a corpus of images with no caption.}
    \label{fig:pilot-T2I-Results}
\end{figure}

To investigate the impact of visual modality collapse on retrieval performance, we evaluated the Recall@20 (R@20) and Recall@100 (R@100) metrics for both \Marvel{} and \UniVLDR{} in the T2I task. As shown in Figure \ref{fig:pilot-T2I-Results}, when image documents lack captions, \Marvel{} achieves an R@100 of only 9.52\%, which is significantly lower compared to the 34.77 achieved by \UniVLDR{}.

\subsection{Semantic Misalignment Issue}
Although \UniVLDR{} is less prone to the visual modality collapse issue compared to \Marvel{}, it is more affected by the semantic misalignment problem, where semantically similar content is positioned far apart in the embedding space. We analyze this phenomenon in this section.

\begin{figure}[]
    \centering
    \begin{subfigure}{0.23\textwidth}
        \centering
        \includegraphics[width=\textwidth]{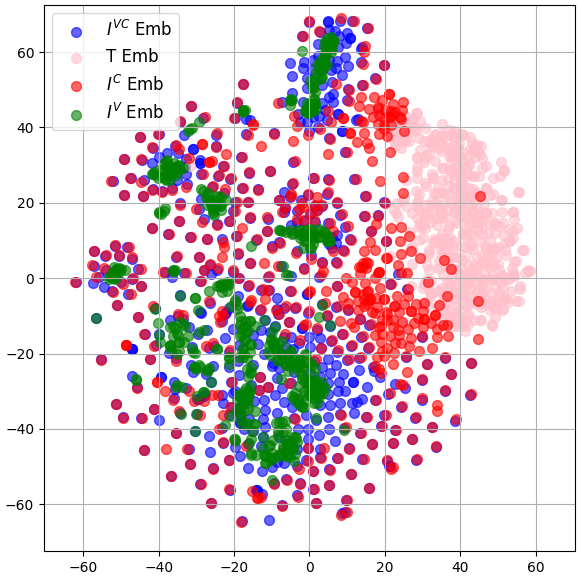}
        \caption{Docs in Marvel space}
        \label{fig:c}
    \end{subfigure}
    \hspace{1mm} 
    \begin{subfigure}{0.23\textwidth}
        \centering
        \includegraphics[width=\textwidth]{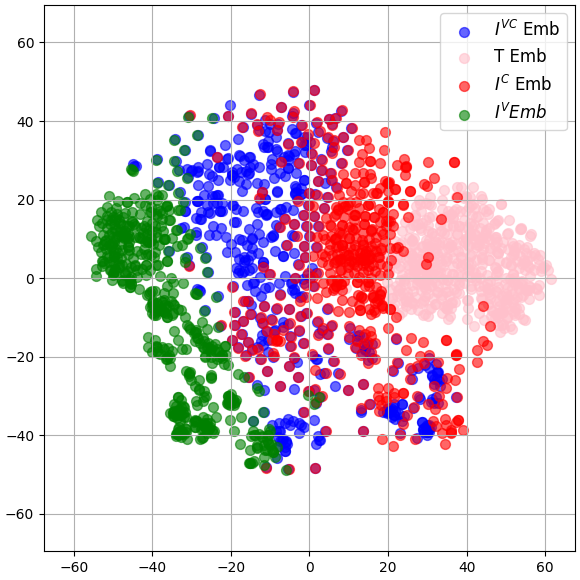}
        \caption{Docs in \UniVLDR{} space}
        \label{fig:d}
    \end{subfigure}

    \caption{Embeddings of documents with different modalities, where $\imgVC{}$ indicates fused embeddings from both visual and captions.}
    \label{fig:total_space}
\end{figure}

\paragraph{Experimental Setup} To analyze the alignment of document embeddings across different modalities, we utilized the models trained in Section \ref{sec:3.1} to extract representations for four distinct document types: $\imgV{}$, $\imgVC{}$, $\imgC{}$, and $\textDoc{}$. In this context, $\textDoc{}$ refers to textual documents, while $\imgV{}$ and $\imgC{}$ represent the separate visual and caption-based embeddings associated with a single image. Furthermore, $\imgVC{}$ denotes the fused multimodal embeddings derived from both visual and textual information. We then employ t-SNE to map the embeddings for visualization as shown in Figure \ref{fig:total_space}.

\begin{figure}
    \centering
    \includegraphics[width=0.87\linewidth]{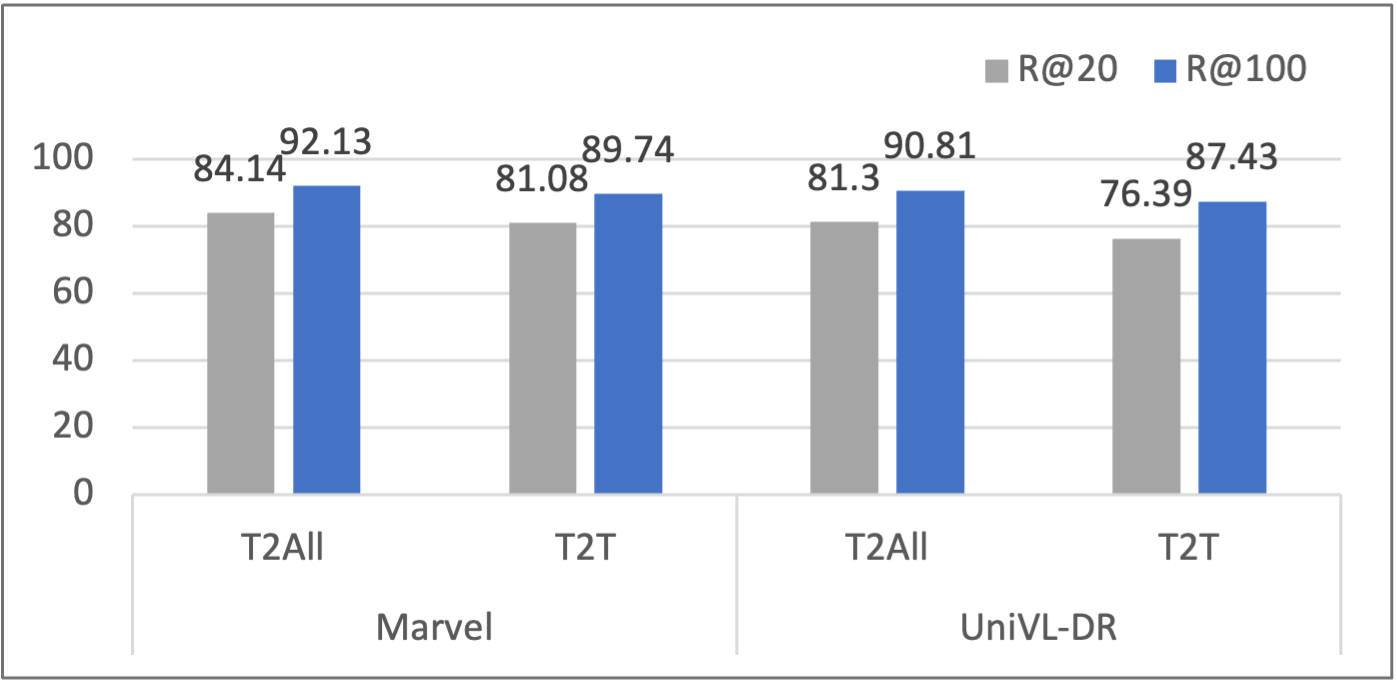}
    \caption{T2T and T2ALL using \Marvel{} and \UniVLDR{}: T2ALL retrieves from a corpus of $\imgVC{}$ and $\textDoc{}$ Docs; T2T retrieves from  a corpus of $\textDoc{}$ Docs.}
    \label{fig:pilot-T2T-Results}
\end{figure}

\paragraph{Results and Discussion} From Figure \ref{fig:total_space}, it is observable that, despite the collapse of the visual modality in the Marvel embedding space, $\imgVC{}$ representations remain closely aligned with their corresponding $\imgV{}$ and $\imgC{}$ representations. In contrast, representations of $\imgC{}$, $\imgV{}$ and $\imgVC{}$ in the \UniVLDR{} space tend to cluster separately. There is a gap between different modal representations of the same semantic entity.
Table \ref{tab:cossim} also shows that in \UniVLDR{}, the average cross-modal similarity s($\imgV{}$,$\imgC{}$) for the same image is 0.27, notably lower than the 0.491 average similarity between image captions and unrelated texts, where the most dissimilar text document is selected for each image as the comparison baseline.

The semantic misalignment issue associated with \UniVLDR{} leads to its lower overall performance and T2T performance compared to \Marvel{} on the WebQA dataset, as shown in Figure \ref{fig:pilot-T2T-Results}.

\begin{table}[!]
    \centering
    \resizebox{0.48\textwidth}{!}{%
        \begin{tabular}{c|ccccc} 
        \toprule
        Method & s($\imgV{}$,$\textDoc{}$) & s($\imgV{}$,$\imgC{}$) & s($\imgC{}$,$\textDoc{}$) & s($\imgVC{}$,$\textDoc{}$) \\
        \midrule
        
        \textbf{\Marvel{}} & 0.103 & 0.494 & 0.206 & 0.098 \\
        \midrule
        \textbf{\UniVLDR{}} & 0.073 & \todo{0.270} & \textbf{0.491} & \textbf{0.391} \\
            \bottomrule
        \end{tabular}
    }
    \caption{Cosine similarity between embeddings of different modalities of WebQA+ Dataset.}
    \label{tab:cossim}
\end{table}

\section{Our Methodology}



\begin{figure}[]
    \centering
    \begin{subfigure}{0.23\textwidth}
        \centering
        \includegraphics[width=\textwidth]{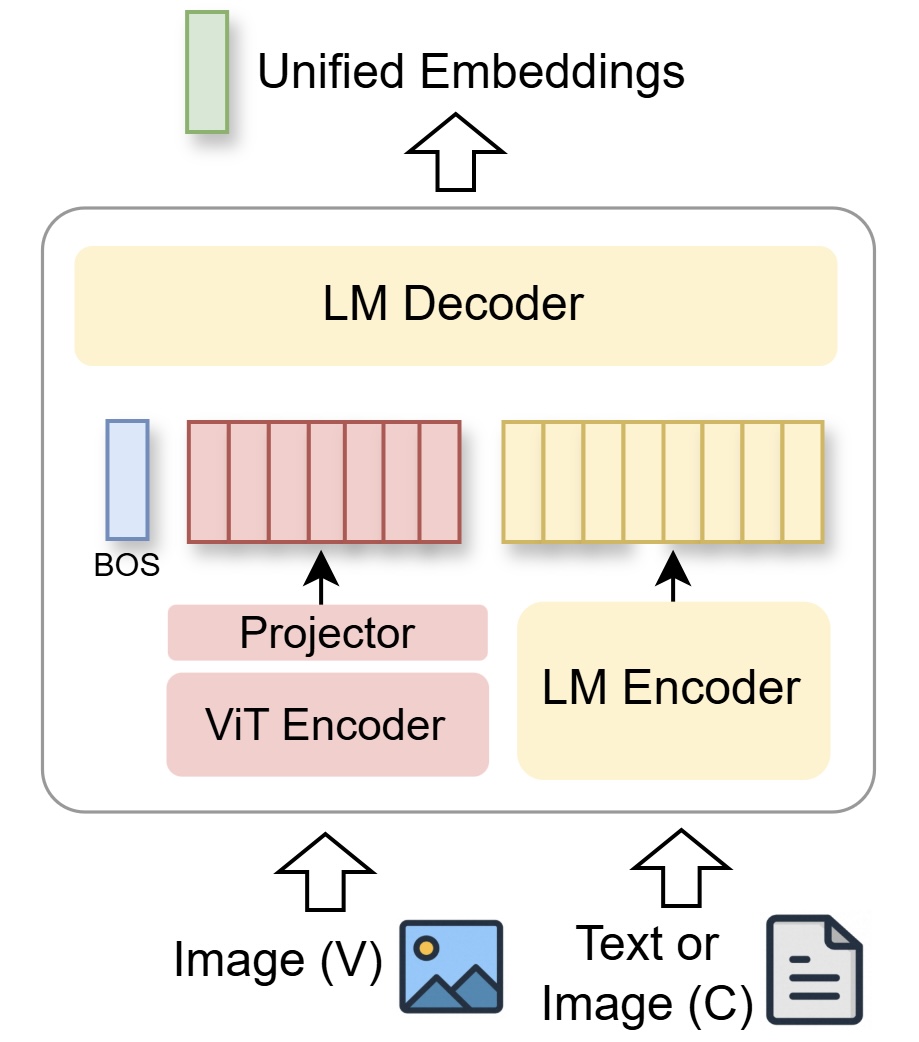}
        \caption{FiD Architecture.}
        \label{fig:mimic-architecture}
    \end{subfigure}
    \hspace{1mm} 
    \begin{subfigure}{0.23\textwidth}
        \centering
        \includegraphics[width=\textwidth]{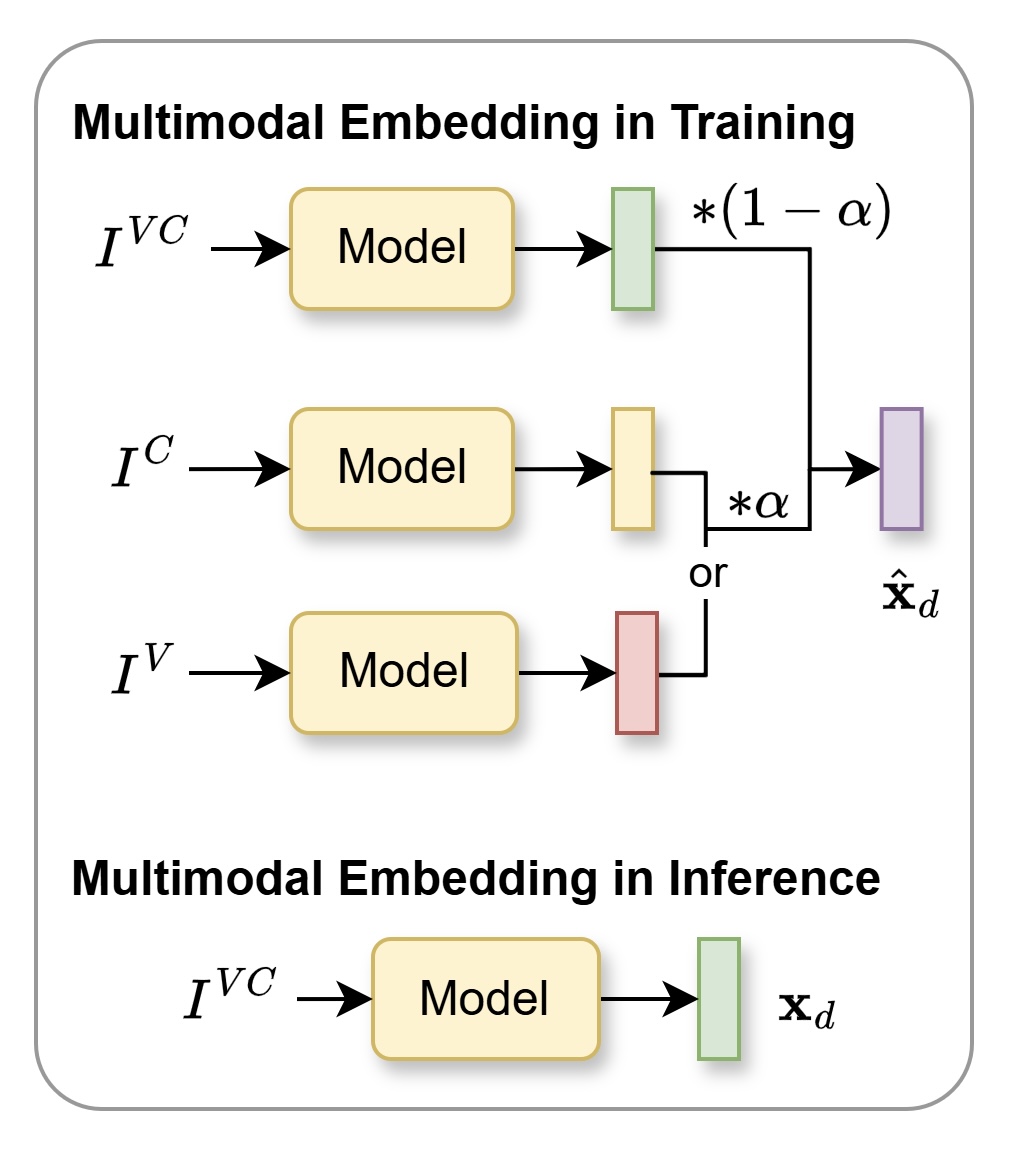}
        \caption{Single-Modality MixIn.}
        \label{fig:mimic-training}
    \end{subfigure}

    \caption{(left) The encoding of different modalities under our architecture. (right) The multi-modal representation in training and inference phase.}
  \label{fig:our_archi}
\end{figure}

\paragraph{\ourmethod{} Architecture}  \ourmethod{} uses separate encoders, like late-fusion models, to retain modality-specific features. However, rather than relying on addition to fuse, we introduce cross-attention in an LM decoder to aggregate representations (i.e., Fusion-in-decoder or FiD) and enhance semantic alignment to match that of the early-fusion approach. Our architecture is demonstrated in Figure \ref{fig:mimic-architecture}. The framework supports encoder–decoder models (e.g., BART \cite{lewis2020bart}, T5, PaLI-X \cite{chen2023palixscalingmultilingualvision}, T5Gemma \cite{zhang2025encoderdecodergemmaimprovingqualityefficiency}) and can pair decoder-only LLMs with LLM encoders (e.g., LLM2Vec \cite{behnamghader2024llmvec}, Qwen3Embedding \cite{zhang2025qwen3embeddingadvancingtext}, LLM2Comp \cite{zhang2025learningcompressunlockingpotential}). Here, we employ T5 as the LM and CLIP as the vision encoder. Exploration with large VLM-based extensions is left for future work.

More formally, our task is to represent multimodal documents from a multimodal corpus $\mathcal{D}$ for retrieval, where each document $I^{VC}_d\in \mathcal{D}$ is a multi-modal document with textual and visual modalities in the general case. In practice, one of the modalities of the document can be missing, in this case, $T_d$ denotes a text-only document, and $I^{V}_d$ denotes a document with only visual information. Similarly, on the query side, we denote $I_q^{VC}$ as multimodal queries, $T_q$ as text-only queries. For a multimodal document, the embedding is obtained by passing both modalities to \ourmethod{} as shown in Figure \ref{fig:mimic-architecture}. The visual and text features serve as key–value vectors, and the [BOS] token in the T5 decoder acts as the query to generate the representation. There is no self-attention interaction between text and visual features, and the fusion occurs only through cross-attention in the decoder.

The embedding $\mathbf{x}_d$ is obtained as follows:
\begin{equation}
\begin{aligned}
e_d^{T} =& \text{LM-Encoder}(I^{C}_d) \\[1.5ex]
e_d^{V} =& \text{Projector}(\text{Visual-Encoder}(I^{V}_d)) \\[1.5ex]
\mathbf{x}_d =& \text{LM-Decoder}(\text{Concat}(e_d^{V},e_d^{T}))
\end{aligned}
\label{eq:archi-coding}
\end{equation}
where Visual-Encoder is the image encoder from the CLIP model, LM-Encoder and LM-Decoder indicate the T5 encoder and decoder, respectively. For each multi-modal document, we can also obtain the single modal representations: $\mathbf{x}^V_d$, visual representation of the $d$ document, and $\mathbf{x}^T_d$, the textual representation of the $d$ document, by passing the single modality to both the LM encoder and decoder. Specifically, we have:
\begin{equation}
\begin{aligned}
\mathbf{x}^T_d =& \text{LM-Decoder}(e_d^{T}) \\[1.5ex]
\mathbf{x}^V_d =& \text{LM-Decoder}(e_d^{V})
\end{aligned}
\label{eq:archi-coding-single-modality}
\end{equation}
For full-text document $T_d$, we obtain the representation $\mathbf{x}_d=\mathbf{x}^T_d$, where $\mathbf{x}^T_d$ is decoded as in Eq. \ref{eq:archi-coding-single-modality}. Similarly, the visual-only document has the representation $\mathbf{x}_d=\mathbf{x}^V_d$. We obtain multi-modal query representation $\mathbf{x}_q$ in the same way as for encoding candidate documents, using the same network model and parameters. 

\paragraph{Training} Contrastive learning is employed to train \ourmethod{} so that a positive pair of $(\mathbf{x}_q^i,\mathbf{x}_d^i)$ should be close to each other, whereas a negative pair $(\mathbf{x}_q^i,\mathbf{x}_d^j)$ is far apart, and the loss is shown in Eq. \ref{eq:contrastive-loss}. 
Here, the positive pair means the pair of the (groundtruth) relevant document with the query, and the negative pair is sampled from the set of groundtruth documents of other queries in the batch (in-batch negative). 
A balanced number of samples of different modalities $(\mathbf{x}_q,\mathbf{x}_d)$ pairs is supported by the dataset, as shown in Section \ref{sec:Setup}.
\begin{equation}
\begin{aligned}
    L &= -\frac{1}{N} \sum_{i=1}^{N} \log \frac{\exp(f(\mathbf{x}_q^i, \mathbf{x}_d^{i}))}{\sum_{j=1}^{N} \exp(f(\mathbf{x}_q^i, \mathbf{x}_d^j))} 
\end{aligned}
\label{eq:contrastive-loss}
\end{equation}



\paragraph{Single-Modality Mixin} During the training phase, for multimodal documents, we mix the single modality $\mathbf{x}^V_d$ and $\mathbf{x}^T_d$ representations (see Eq. \ref{eq:archi-coding-single-modality}) into the fused representation $\mathbf{x}_d$, and obtain mix-in representation as follows: \begin{equation}
\begin{aligned}
\hat{\mathbf{x}}_{d} &= (1-\alpha)\mathbf{x}_d\\
&+\alpha(\delta\times\mathbf{x}^{V}_d+(1-\delta)\times\mathbf{x}^{T}_d) 
\label{eq:guide}
\end{aligned}
\end{equation}
where $\alpha$ is randomly sampled from $[0,\bar{\alpha}]$ ($\bar{\alpha}<1$) to adjust the contribution of single-modal mixin representation, and $\delta$ is also randomly sampled from $\{0,1\}$ to select which modality to mixin. In a similar way, we can obtain mixin representation for multi-modal queries $\hat{\mathbf{x}}_q$. The single-modal documents and queries are decoded in the same way as previously described. For training with single-modality mixin, we use mixin representations $\hat{\mathbf{x}}_d$, $\hat{\mathbf{x}}_q$ in place of $\mathbf{x}_d$, $\mathbf{x}_q$ in Eq. \ref{eq:contrastive-loss}.

\paragraph{Caption Dropout} Through single-modality mixin, we allow the visual encoder to receive more training signals (gradients) via the mixin representation. 
However, one side of the datasets (either queries or documents) is still purely text-based, and text still dominates. So, it is essential that we actively drop captions from images to force the model to rely on visual information for the retrieval. We maintain the \textit{caption ratio} as a training hyperparameter.


\paragraph{ANCE Training} Like \Marvel{} and \UniVLDR{}, after the first training stage with in-batch negative, we conduct the second training stage, where we adopt hard negative sampling as in ANCE \cite{ANCE}. We use the checkpoint obtained from the first stage of training as the retrieval model to encode the multi-modal corpus and retrieve hard negative examples. Besides the hard-negative sampling, both stages employ single-modality mixin and caption dropout strategies with the same training parameter settings.

\section{Experiment}

\subsection{Experimental Setup}
\label{sec:Setup}

\begin{table}[]
	\centering
    \resizebox{\linewidth}{!}{
	\begin{tabular}{p{1.5cm}|ccc|c}
    \toprule
        \multirow{2}{*}{\textbf{Dataset}} & \multicolumn{3}{c|}{\textbf{Query}} & \multirow{2}{*}{\textbf{Corpus}} \\

     & \textbf{Train} & \textbf{Test} & \textbf{Valid} & \\
         \midrule

         \multirow{2}{*}{\textbf{EVQA+}} & T\ /\ TI & T\ /\ TI & T\ /\ TI & T\\
         & 15,366\ /\ 32,819 & 500\ /\ 1,049 & 2,455\ /\ 3,750 & 839,692\\
         \midrule
         \multirow{2}{*}{\textbf{WebQA+}} & T & T & T & I\ /\ T\\
         & 31,766 & 5,000 & 4,966  & 389,750\ /\ 787,697\\

    \bottomrule
\end{tabular}}
 \caption{Dataset Statistics. On the query side, ``T'' denotes the text modality question and ``TI'' represents a multimodal question composed of text and image. On the document side, ``I'' refers to image documents, and ``T'' refers to text documents. We retain or discard captions of queries or documents that include images based on the caption ratio.}
  \label{tab:dataset-statistics}  
\end{table}

\begin{table*}[t!]
  \centering
  \resizebox{1.0\textwidth}{!}{
  \begin{tabular}{c|c|l|cccccccc}
    \toprule   
    \textbf{Dataset} & \textbf{Task} & \textbf{Method} & \textbf{R@1} & \textbf{R@5} & \textbf{R@20} & \textbf{R@100} & \textbf{MRR@10}  & \textbf{NDCG@10} & \textbf{MRR@20}& \textbf{NDCG@20} \\
    \midrule
    \multirow{7}{*}{\textbf{WebQA+}} 

     & \multirow{7}{*}{\texttt{T->All}}     

    &  \Marvel{}-DPR &32.54	&43.63	&58.02	&68.07	&43.89	&40.89	&44.19	&42.93   \\
        & & \Marvel{}-ANCE &40.84	&53.13	&65.33	&72.10	&52.61	&49.48	&52.78	&51.00 \\

     & & VISTA*  &21.73	&31.59	&43.57	&52.66	&31.63	&29.31	&31.81	&30.86 \\

    & & CLIP-DPR &28.65	&39.00	&54.16	&66.56	&39.80	&36.69	&40.14	&38.78  \\
    & & \UniVLDR{} &36.81	&49.55	&62.39	&69.65	&49.21	&46.27	&49.40	&47.75  \\
    \rowcolor{lightgray}
     \cellcolor{white} & \cellcolor{white} & MiMIC &32.62 &44.35	&60.29	&73.07	&44.62	&41.75	&44.96	&43.88   \\
     \rowcolor{lightgray}
     \cellcolor{white} & \cellcolor{white} & MiMIC-ANCE & 40.96 & 54.88 & 69.67 & 78.87 & 54.11 & 50.97 & 54.38 & 52.77  \\
    \midrule

    \multirow{7}{*}{\textbf{EVQA+}} 

     & \multirow{7}{*}{\texttt{ALL->T}} 

    &  \Marvel{}-DPR & 26.17	&37.92	&49.75	&59.55	&35.74	&34.74	&36.02	&36.3\\
    & & \Marvel{}-ANCE &31.85	&45.26	&57.51	&66.8	&42.26	&41.5	&42.5	&42.96 \\
    
    & & VISTA* & 24.05	&36.16	&49.46	&60.27	&33.78	&33.05	&34.05	&34.73 \\

    & & CLIP-DPR & 27.86	&41.19	&55.32	&68.34	&38.38	&37.62	&38.72	&39.39 \\
    & & \UniVLDR{} & 30.48	&45.35	&58.9	&70.75	&41.58	&41.02	&41.89	&42.66\\

    \rowcolor{lightgray}
    \cellcolor{white} & \cellcolor{white} & MiMIC  & 29.41	&43.08	&57.06	&71.13	&40.03	&39.30	&40.36	&41.15 \\

    \rowcolor{lightgray}
    \cellcolor{white} & \cellcolor{white} & MiMIC-ANCE  & 33.36 & 47.90 & 61.28 & 73.87 & 44.21 & 43.64 & 44.51 & 45.28 \\   
    \bottomrule
  \end{tabular}
  }
  \caption{Retrieval Performance on WebQA+ (based on a mixed-modal corpus $\corpusall{}$) and EVQA+.}    
  \label{tab:mainresult}
\end{table*}

\paragraph{Dataset} WebQA \cite{WebQA} is a multi-hop, multi-modal retrieval benchmark, including T2T and T2I retrieval tasks, with a task ratio of the two tasks is 1:1. Encyclopedic-VQA \cite{mensink23iccv} is a visual question answering (VQA) dataset, and the questions are designed for multi-modal question answering. See Appendix \ref{sec:appendix-data} for more details.
For our experiments, we extend the WebQA and EVQA datasets to WebQA+ and EVQA+, respectively. The dataset statistics are shown in Table \ref{tab:dataset-statistics}. 

In WebQA+, aside from 50\% of the images in its corpus lacking captions according to a caption ratio of 0.5, everything else is consistent with WebQA, and the search settings still include T2T and T2I tasks. EVQA+ is obtained by sampling the original EVQA dataset and mixing it with WebQA T2T data, the retrieval settings include T2T and TI2T tasks, where the introduction of the T2T task enables the dataset to support a wider range of multimodal retrieval scenarios. In the TI2T setting of EVQA+, each query consists of a question (text) and an image, where the image also retains captions based on a caption ratio of 0.5. 

\paragraph{Baselines}
We compare our results with some baseline methods: \UniVLDR{}, \Marvel{}, VISTA. Here, CLIP-DPR and Marvel-DPR respectively are \UniVLDR{} and \Marvel{}-ANCE without the second stage training (ANCE training with hard negatives). The training details of baselines and \ourmethod{} are given in Appendix \ref{sec:appendix1}, \ref{sec:appendix2}.




\paragraph{Metrics} We evaluate retrieval performance using Recall@1/5/20/100, MRR@10/20, and NDCG@10/20.
Recall@k measures the proportion of relevant items retrieved within the top-k results. NDCG@k evaluates ranking quality by accounting for relevance and position. MRR@k reflects the average reciprocal rank of the first relevant item. 



\subsection{Main Results} 
Table \ref{tab:mainresult} presents the overall retrieval performance across various multimodal retrieval tasks on WebQA+ and EVQA+.
Among the baselines without ANCE training, \Marvel{}-DPR shows a clear advantage over CLIP-DPR on WebQA+ (e.g., +3.89 R@1). On EVQA+, CLIP-DPR proves more robust, outperforming \Marvel{}-DPR by 1.69 in R@1 and 8.79 in R@100. This indicates that EVQA+ places greater emphasis on image understanding, where \Marvel{} is limited due to visual modality collapse in semantic space.

Among the baselines with ANCE training, \Marvel{}-ANCE is the strongest baseline on WebQA+, surpassing \UniVLDR{} by 4.03 in R@1. On EVQA+, it remains competitive but lags behind \UniVLDR{} in R@20 and R@100 (–3.95 in R@100). While \Marvel{} struggles with processing visual information, its semantic space representation is still better than \UniVLDR{}.

For VISTA, we adopted CLIP-B-32 as the image encoder in order to be consistent with other baselines, we denote this variant as VISTA* for reference, while the original VISTA used EVA-CLIP. VISTA* performs poorly on both datasets, showing that it is less robust in missing-modality situations.

\ourmethod{} shows superior performance across both evaluated scenarios. In the base configuration without ANCE training, \ourmethod{} outperforms \Marvel{}-DPR, CLIP-DPR, and the VISTA baseline. In particular, in several key metrics, \ourmethod{} even exceeds \Marvel{}-ANCE and \UniVLDR{}, achieving an R@100 of 73.07 on WebQA+, compared to 69.65 for \UniVLDR{} and 72.10 for \Marvel{}-ANCE.
\textit{When integrated with the ANCE strategy, \ourmethod{}-ANCE attains state-of-the-art performance across all categories}. Compared with the strongest baseline, \Marvel{}-ANCE, our \ourmethod{}-ANCE improves R@100 from 72.10 to 78.87 on WebQA+ (a 6.77 absolute gain) and from 66.80 to 73.87 on EVQA+ (a 7.07 absolute gain).



\subsection{Different Multi-modal Retrieval Settings} 
\begin{figure}[t]
  \centering
  \includegraphics[width=\linewidth]{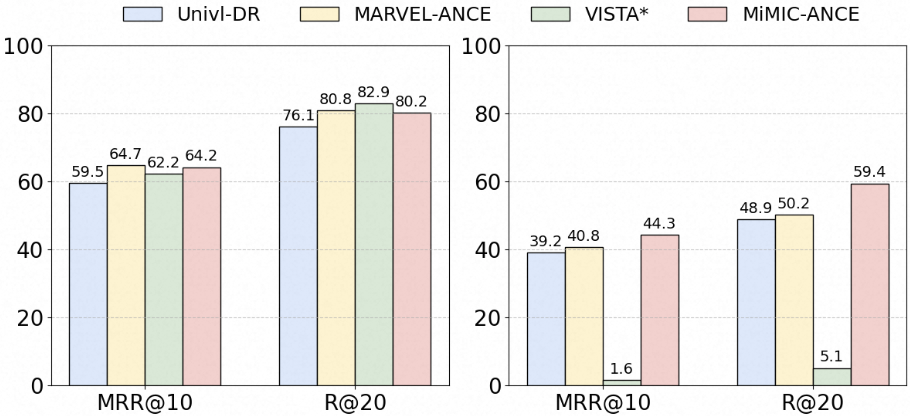}
  \caption{Different modality retrieval tasks: T2T (left) and T2I (right) on WebQA+. All tasks are retrieved in a mixed Corpus $\corpusall{}$ but reported separately.}
  \label{fig:webqa_single}
\end{figure}

To analyze the contribution of different modalities, we perform evaluation for modality-specific retrieval tasks. For WebQA+, we separately measure T2T and T2I tasks (Figure \ref{fig:webqa_single}). Note that the retrieval is over a corpus with mixed documents, which is different from that in Section \ref{sec:pilot}. The results for EVQA+ are reported in Appendix \ref{sec:dmr-evqa+}.

The results in Figure \ref{fig:webqa_single} show that \textit{\ourmethod{}-ANCE matches strong text retrieval baselines such as \Marvel{}-ANCE and VISTA* in T2T retrieval, while outperforming them substantially in T2I}. In the T2T task, \ourmethod{}-ANCE performs on par with \Marvel{}-ANCE (R@20 = 80.8) and VISTA* (R@20 = 82.9), achieving R@20 = 80.2. In the T2I task, \ourmethod{}-ANCE achieves a clear advantage, reaching R@20 = 59.4, which surpasses \Marvel{}-ANCE (R@20 = 50.2) by +9.2 and \UniVLDR{} (R@20 = 48.9) by +10.5. Notably, while VISTA* excels in text retrieval, its performance drops sharply in T2I (R@20 = 5.1), revealing severe modality bias. In contrast, \ourmethod{}-ANCE maintains balanced and robust performance across modalities, consistent with the observations in Table \ref{tab:mainresult}.



\subsection{Ablation Study}

\begin{table}[!]
    \centering
    \resizebox{0.48\textwidth}{!}{%
        \begin{tabular}{lccc}
        \toprule
        \textbf{MiMIC} & \textbf{MRR@10} & \textbf{Recall@20} & \textbf{Recall@100} \\
        \midrule
         \cellcolor{gray!15}On WebQA+ Dataset& \cellcolor{gray!15}\textbf{44.62} & \cellcolor{gray!15}\textbf{60.29} & \cellcolor{gray!15}\textbf{73.07} \\
          
          \quad w/o Caption Dropout & 44.05 & 58.05 & 67.77 \\
          \quad w/o Single-Modality MixIn & 44.57 & 60.21 & 72.99 \\
          \quad w/ Early-Fusion (\Marvel{}-DPR) & 44.22 & 58.59 & 71.84 \\
          
          \quad w/ Late-Fusion (CLIP-DPR) &  37.25 & 54.27 & 71.81\\
          
        \cmidrule{1-4}
        
          \cellcolor{gray!15}On EVQA+ Dataset & \cellcolor{gray!15}\textbf{40.03} & \cellcolor{gray!15}\textbf{57.06} & \cellcolor{gray!15}\textbf{71.13} \\
          
          \quad w/o Caption Dropout & 38.58 & 52.90 & 63.61 \\

        \quad w/o Single-Modality MixIn & 38.19 & 54.57 & 68.51 \\
                    
          \quad w/ Early-Fusion (\Marvel{}-DPR) & 38.01 & 54.09 & 68.33 \\
          
          \quad w/ Late-Fusion (CLIP-DPR) & 36.87  & 54.01 & 69.10 \\
          
        \bottomrule
        \end{tabular}
    }
    \caption{Ablation Study for MiMIC on WebQA+ and EVQA+ Datasets. (without ANCE training). }
    \label{tab:abl1}
\end{table}

The ablation study examines the contribution of each component in \ourmethod{} (without ANCE training). We evaluate performance changes under four settings: (1) w/o Caption Dropout and (2) w/o Single-Modality Mixin, where each training strategy is removed in turn; (3) w/ Early Fusion (\Marvel{}), using the \Marvel{}-DPR architecture with \ourmethod{}’s training strategies without ANCE training; and (4) w/ Late Fusion, using the CLIP-DPR architecture (which is also the same as \UniVLDR{}) with the same training strategies. The second and third variants assess the effect of the fusion method.

The results in Table \ref{tab:abl1} indicate that caption dropout is essential for both datasets, leading to substantial declines in R@20 ($-2.24$ for WebQA+ and $-4.16$ for EVQA+) and R@100 (-5.30 for WebQA+ and -7.52 for EVQA+) when removed. In contrast, single-modality mixin has a stronger effect on EVQA+ where removing this strategy causes a 1.60-point drop in R@100, compared to only 0.06 on WebQA+.

By comparing \ourmethod{} with variants where the fusion-in-decoder design is replaced by early fusion (as in \Marvel{}-DPR) or late fusion (as in CLIP-DPR), we observe noticeable drops in all metrics across both datasets. These results confirm the effectiveness of the FiD architecture in maintaining balanced multimodal retrieval performance.

\subsection{Further Analysis}

\begin{figure}[]
    \centering
    \begin{subfigure}{0.23\textwidth}
        \centering
        \includegraphics[width=\textwidth]{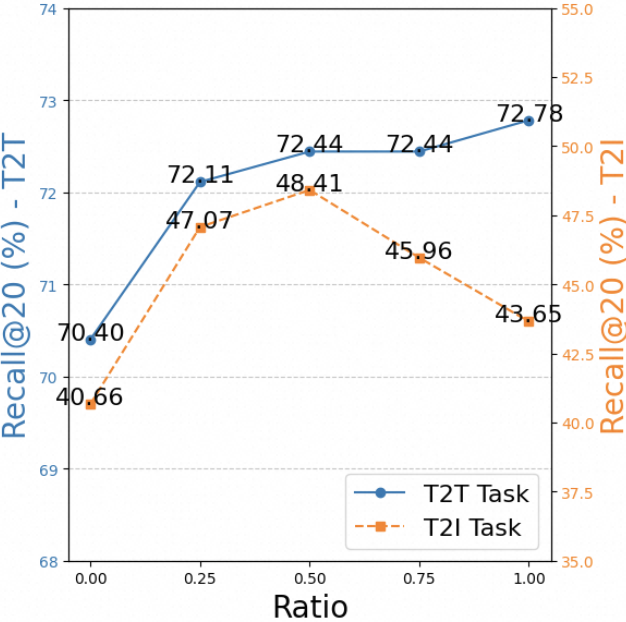}
        \caption{on WebQA+ Datasets}
        \label{fig:ratio-webqa}
    \end{subfigure}
    \hspace{1mm} 
    \begin{subfigure}{0.23\textwidth}
        \centering
        \includegraphics[width=\textwidth]{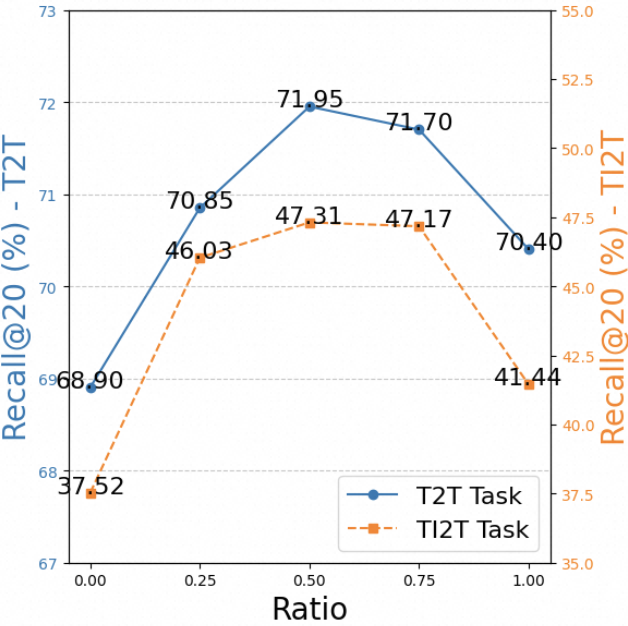}
        \caption{on EVQA+ Datasets}
        \label{fig:ratio-evqa}
    \end{subfigure}

    \caption{The Recall@20 performance of MiMIC using different Caption Ratios.}
    \label{fig:ratio}
\end{figure}

\paragraph{The Impact of Caption Ratio} We vary the caption dropout ratio during training and evaluate performance on T2T and T2I tasks in WebQA+, as well as T2T and TI2T tasks in EVQA+. As shown in Figure \ref{fig:ratio}, the best results for T2I and TI2T tasks are achieved with caption ratios between 0.25 and 0.75. Interestingly, both extreme settings—0 (no captions) and 1.0 (all captions used)—degrade performance, suggesting that captions may act as a bridge, helping to map the image to semantic space.

\begin{figure}[]
    \centering
    \begin{subfigure}{0.23\textwidth}
        \centering
        \includegraphics[width=\textwidth]{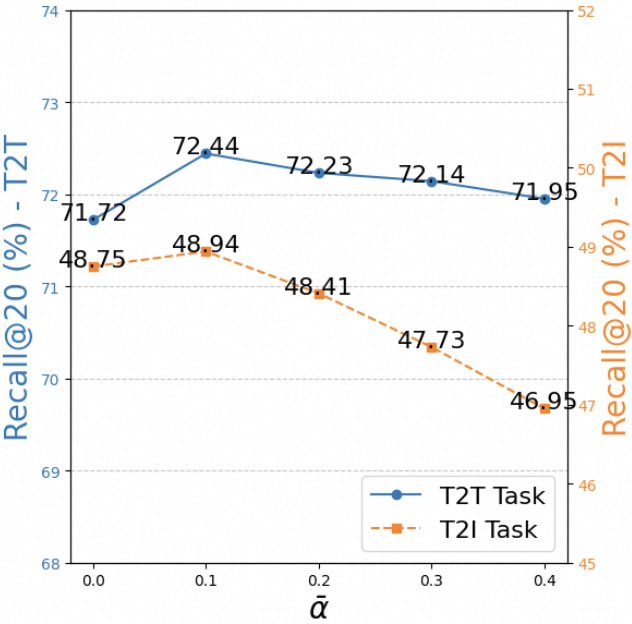}
        \caption{on WebQA+ Datasets}
        \label{fig:alpha-webqa}
    \end{subfigure}
    \hspace{1mm} 
    \begin{subfigure}{0.23\textwidth}
        \centering
        \includegraphics[width=\textwidth]{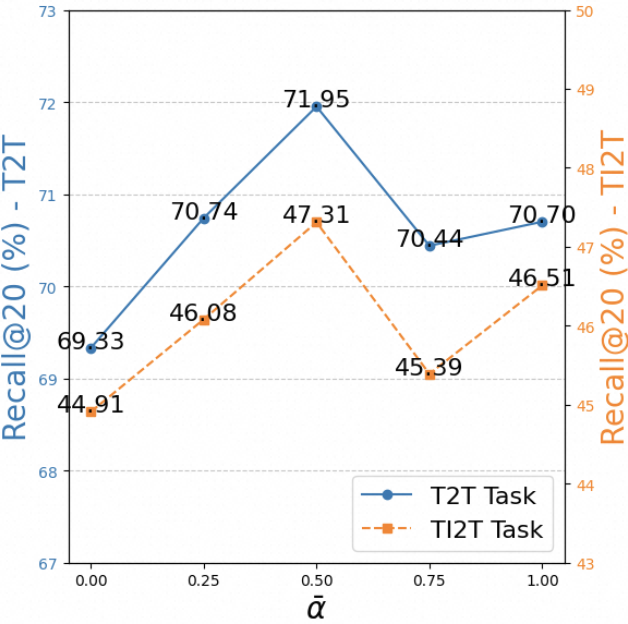}
        \caption{on EVQA+ Datasets}
        \label{fig:alpha-evqa}
    \end{subfigure}

    \caption{The Recall@20 performance varies with the increasing $\bar{\alpha}$ in Single-Modality MixIn of MiMIC.}
    \label{fig:abl-scale}
\end{figure}

\paragraph{The Impact of Single-Modality MixIn} To examine the effect of modality mixin, we vary the upper bound of the mix-in ratio, $\bar{\alpha}$. As shown in Figure \ref{fig:abl-scale}, the influence of single-modality mix-in appears dataset-dependent: the optimal ratio is around 0.0–0.2 for WebQA+, and approximately 0.5 for EVQA+. For datasets that rely more on vision when understanding semantics, performance improves more markedly as $\bar{\alpha}$ increases. However, excessive increases introduce more noise and lead to performance degradation.

\paragraph{Embedding Space of \ourmethod{}} 

\begin{figure}[t]
    \centering
    \begin{subfigure}{0.23\textwidth}
        \centering
        \includegraphics[width=\textwidth]{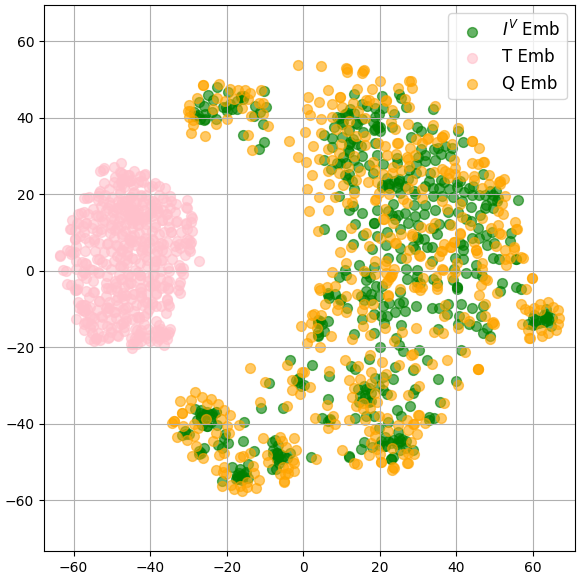}
        \caption{MiMIC: $Q \rightarrow\corpusIv{}$}
        \label{fig:our_space-a}
    \end{subfigure}
    \hspace{1mm} 
    \begin{subfigure}{0.23\textwidth}
        \centering
        \includegraphics[width=\textwidth]{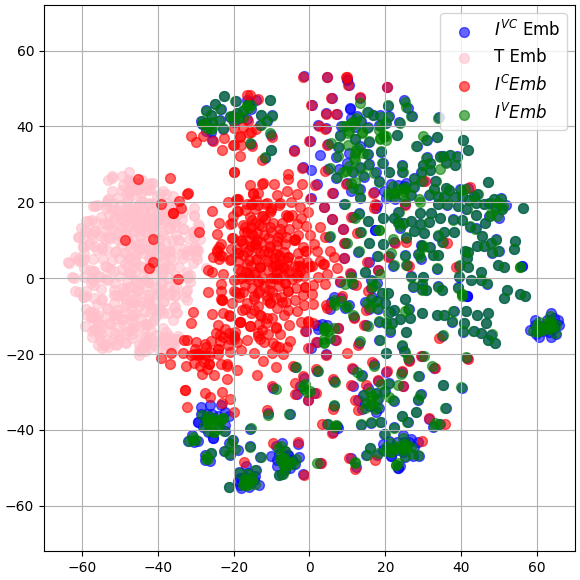}
        \caption{Docs in MiMIC space}
        \label{fig:our_space-b}
    \end{subfigure}

    \caption{Embeddings in MiMIC Space}
    \label{fig:our_space}
\end{figure}

Consistent with Section \ref{sec:pilot}, we visualized the representations from \ourmethod{}. As shown in Figure \ref{fig:our_space-a}, the query embeddings align effectively with their corresponding image document embeddings $I^V$, even in missing-modality situations. Furthermore, the image representations exhibit a diverse and well-distributed pattern without obvious representation collapse. 
Figure \ref{fig:our_space-b} demonstrates that $I^V$ and $I^{VC}$ exhibit consistent distributions with no significant modality gap. The caption embeddings $I^C$ serve as a semantic bridge: some captions surround the image, while others lie somewhere between the text and image modalities.

To further evaluate the semantic alignment of \ourmethod{}, we calculate the neighborhood overlap across modalities to measure semantic misalignment, expecting that the bigger overlap leads to better alignment. Specifically, we first sample a total of 500 sample points of WebQA+ dataset, then calculated how many of the k nearest neighbors of $I^V$ (image visual content) and 
$I^C$ (image captions) are the same points. Table \ref{tab:misalignment} shows the neighborhood overlap for different models for $k=5$ and $k=50$, demonstrating that our model has the best semantic alignment compared to \UniVLDR{} and \Marvel{}-ANCE again.

\begin{table}[!]
    \centering
    \resizebox{0.45\textwidth}{!}{%
    \begin{tabular}{lccccccc}
    \toprule
    Model & & & k=5 & & & k=50 & \\
    \midrule
    \UniVLDR{} & & & 0.0204 & & & 0.0182 & \\
    \Marvel{}-ANCE & & & 0.0230 & & & 0.0208 & \\
    \textbf{\ourmethod{}-ANCE} & & & \textbf{0.1010} & & & \textbf{0.0936} & \\
    \bottomrule
    \end{tabular}
    }
    \caption{Measuring semantic misalignment with cross-modality neighborhood overlap.}
    \label{tab:misalignment}
\end{table}

\section{Conclusion}

This work identifies and addresses two primary failure modes in UMR: visual modality collapse and semantic misalignment. By introducing Fusion-in-Decoder (FiD), we employ language models and cross-attention to dynamically integrate multimodal signals. Combined with our robust training strategy—incorporating single-modality mix-in and caption dropout—our approach significantly outperforms strong baselines on WebQA+ and EVQA+, demonstrating particular resilience in incomplete data scenarios.


\section*{Limitations}

Our work offers pioneering insights into the visual collapse and semantic alignment issues of multimodal representation spaces in UMR. However, current research has not yet balanced the performance of various multimodal tasks. Our work is also a preliminary exploration, primarily motivated by observations on missing-modality WebQA+ and EVQA+ datasets. Future research will investigate the impact of datasets with different modality types on shaping the multimodal representation space, and extend our work to larger models trained on more data. Furthermore, a deeper investigation into dynamically adjusting $\bar\alpha$ in Single-Modality MixIn is left for future work, as different training stages may benefit from different mix-in levels.
There is still considerable room for research on how multimodal representations can fully exploit information across modalities, and on the interplay between different modalitiess in the UMR task.

\section*{Acknowledgments}

We thank the anonymous reviewers and area chair for their insightful feedback. This work was supported by the National Natural Science Foundation of China  (Grant No. U2342218, W2532049) and the Fundamental and Interdisciplinary Disciplines Breakthrough Plan of the Ministry of Education of China No. JYB2025XDXM118.

\bibliography{custom}

\begin{thebibliography}{35}
\providecommand{\natexlab}[1]{#1}

\bibitem[{BehnamGhader et~al.(2024)BehnamGhader, Adlakha, Mosbach, Bahdanau, Chapados, and Reddy}]{behnamghader2024llmvec}
Parishad BehnamGhader, Vaibhav Adlakha, Marius Mosbach, Dzmitry Bahdanau, Nicolas Chapados, and Siva Reddy. 2024.
\newblock \href {https://openreview.net/forum?id=IW1PR7vEBf} {{LLM}2vec: Large language models are secretly powerful text encoders}.
\newblock In \emph{First Conference on Language Modeling}.

\bibitem[{Chang et~al.(2022)Chang, Narang, Suzuki, Cao, Gao, and Bisk}]{WebQA}
Yingshan Chang, Mridu Narang, Hisami Suzuki, Guihong Cao, Jianfeng Gao, and Yonatan Bisk. 2022.
\newblock Webqa: Multihop and multimodal qa.
\newblock In \emph{Proceedings of the IEEE/CVF Conference on Computer Vision and Pattern Recognition (CVPR)}.

\bibitem[{Chaudhuri et~al.(2025)Chaudhuri, Dutta, Bui, and Georgescu}]{chaudhuri2025closer}
Abhra Chaudhuri, Anjan Dutta, Tu~Bui, and Serban Georgescu. 2025.
\newblock A closer look at multimodal representation collapse.
\newblock \emph{arXiv preprint arXiv:2505.22483}.

\bibitem[{Chen et~al.(2023{\natexlab{a}})Chen, Djolonga, Padlewski, Mustafa, Changpinyo, Wu, Ruiz, Goodman, Wang, Tay, Shakeri, Dehghani, Salz, Lucic, Tschannen, Nagrani, Hu, Joshi, Pang, Montgomery, Pietrzyk, Ritter, Piergiovanni, Minderer, Pavetic, Waters, Li, Alabdulmohsin, Beyer, Amelot, Lee, Steiner, Li, Keysers, Arnab, Xu, Rong, Kolesnikov, Seyedhosseini, Angelova, Zhai, Houlsby, and Soricut}]{chen2023palixscalingmultilingualvision}
Xi~Chen, Josip Djolonga, Piotr Padlewski, Basil Mustafa, Soravit Changpinyo, Jialin Wu, Carlos~Riquelme Ruiz, Sebastian Goodman, Xiao Wang, Yi~Tay, Siamak Shakeri, Mostafa Dehghani, Daniel Salz, Mario Lucic, Michael Tschannen, Arsha Nagrani, Hexiang Hu, Mandar Joshi, Bo~Pang, and 24 others. 2023{\natexlab{a}}.
\newblock \href {https://arxiv.org/abs/2305.18565} {Pali-x: On scaling up a multilingual vision and language model}.
\newblock \emph{Preprint}, arXiv:2305.18565.

\bibitem[{Chen et~al.(2023{\natexlab{b}})Chen, Hu, Luan, Sun, Changpinyo, Ritter, and Chang}]{chen-etal-2023-pre-trained}
Yang Chen, Hexiang Hu, Yi~Luan, Haitian Sun, Soravit Changpinyo, Alan Ritter, and Ming-Wei Chang. 2023{\natexlab{b}}.
\newblock \href {https://doi.org/10.18653/v1/2023.emnlp-main.925} {Can pre-trained vision and language models answer visual information-seeking questions?}
\newblock In \emph{Proceedings of the 2023 Conference on Empirical Methods in Natural Language Processing}, pages 14948--14968, Singapore. Association for Computational Linguistics.

\bibitem[{Javaloy et~al.(2022)Javaloy, Meghdadi, and Valera}]{javaloy2022mitigating}
Adri{\'a}n Javaloy, Maryam Meghdadi, and Isabel Valera. 2022.
\newblock Mitigating modality collapse in multimodal vaes via impartial optimization.
\newblock In \emph{International Conference on Machine Learning}, pages 9938--9964. PMLR.

\bibitem[{Karpukhin et~al.(2020)Karpukhin, Oguz, Min, Lewis, Wu, Edunov, Chen, and Yih}]{dpr}
Vladimir Karpukhin, Barlas Oguz, Sewon Min, Patrick Lewis, Ledell Wu, Sergey Edunov, Danqi Chen, and Wen-tau Yih. 2020.
\newblock Dense passage retrieval for open-domain question answering.
\newblock In \emph{Conference on Empirical Methods in Natural Language Processing (EMNLP)}.

\bibitem[{Lerner et~al.(2022)Lerner, Ferret, Guinaudeau, Le~Borgne, Besançon, Moreno, and Lovón~Melgarejo}]{lerner2022viquae}
Paul Lerner, Olivier Ferret, Camille Guinaudeau, Hervé Le~Borgne, Romaric Besançon, Jose~G Moreno, and Jesús Lovón~Melgarejo. 2022.
\newblock \href {https://doi.org/10.1145/3477495.3531753} {{ViQuAE}, a dataset for knowledge-based visual question answering about named entities}.
\newblock In \emph{Proceedings of The 45th International ACM SIGIR Conference on Research and Development in Information Retrieval}, SIGIR’22, New York, NY, USA. Association for Computing Machinery.

\bibitem[{Lewis et~al.(2020)Lewis, Liu, Goyal, Ghazvininejad, Mohamed, Levy, Stoyanov, and Zettlemoyer}]{lewis2020bart}
Mike Lewis, Yinhan Liu, Naman Goyal, Marjan Ghazvininejad, Abdelrahman Mohamed, Omer Levy, Veselin Stoyanov, and Luke Zettlemoyer. 2020.
\newblock Bart: Denoising sequence-to-sequence pre-training for natural language generation, translation, and comprehension.
\newblock In \emph{Proceedings of the 58th annual meeting of the association for computational linguistics}, pages 7871--7880.

\bibitem[{Liaqat et~al.(2025)Liaqat, Nawaz, Zaheer, Saeed, Sajjad, De~Schepper, Nandakumar, Khan, Gallo, and Schedl}]{r2}
Muhammad~Irzam Liaqat, Shah Nawaz, Muhammad~Zaigham Zaheer, Muhammad~Saad Saeed, Hassan Sajjad, Tom De~Schepper, Karthik Nandakumar, Muhammad~Haris Khan, Ignazio Gallo, and Markus Schedl. 2025.
\newblock Chameleon: A multimodal learning framework robust to missing modalities.
\newblock \emph{International Journal of Multimedia Information Retrieval}, 14(2):21.

\bibitem[{Liu et~al.()Liu, Xiong, Lv, Liu, and Yu}]{liuuniversal}
Zhenghao Liu, Chenyan Xiong, Yuanhuiyi Lv, Zhiyuan Liu, and Ge~Yu.
\newblock Universal vision-language dense retrieval: Learning a unified representation space for multi-modal retrieval.

\bibitem[{Luo et~al.(2023)Luo, Fang, Gokhale, Yang, and Baral}]{luo-etal-2023-end}
Man Luo, Zhiyuan Fang, Tejas Gokhale, Yezhou Yang, and Chitta Baral. 2023.
\newblock \href {https://doi.org/10.18653/v1/2023.acl-long.478} {End-to-end knowledge retrieval with multi-modal queries}.
\newblock In \emph{Proceedings of the 61st Annual Meeting of the Association for Computational Linguistics (Volume 1: Long Papers)}, pages 8573--8589, Toronto, Canada. Association for Computational Linguistics.

\bibitem[{Ma et~al.(2022{\natexlab{a}})Ma, Ren, Zhao, Testuggine, and Peng}]{ma2022multimodal}
Mengmeng Ma, Jian Ren, Long Zhao, Davide Testuggine, and Xi~Peng. 2022{\natexlab{a}}.
\newblock Are multimodal transformers robust to missing modality?
\newblock In \emph{Proceedings of the IEEE/CVF conference on computer vision and pattern recognition}, pages 18177--18186.

\bibitem[{Ma et~al.(2022{\natexlab{b}})Ma, Ren, Zhao, Testuggine, and Peng}]{r1}
Mengmeng Ma, Jian Ren, Long Zhao, Davide Testuggine, and Xi~Peng. 2022{\natexlab{b}}.
\newblock Are multimodal transformers robust to missing modality?
\newblock In \emph{Proceedings of the IEEE/CVF conference on computer vision and pattern recognition}, pages 18177--18186.

\bibitem[{Malitesta et~al.(2024)Malitesta, Rossi, Pomo, Di~Noia, and Malliaros}]{r3}
Daniele Malitesta, Emanuele Rossi, Claudio Pomo, Tommaso Di~Noia, and Fragkiskos~D. Malliaros. 2024.
\newblock Do we really need to drop items with missing modalities in multimodal recommendation?
\newblock In \emph{Proceedings of the 33rd ACM International Conference on Information and Knowledge Management}, pages 3943--3948.

\bibitem[{Marino et~al.(2019)Marino, Rastegari, Farhadi, and Mottaghi}]{okvqa}
Kenneth Marino, Mohammad Rastegari, Ali Farhadi, and Roozbeh Mottaghi. 2019.
\newblock Ok-vqa: A visual question answering benchmark requiring external knowledge.
\newblock In \emph{Conference on Computer Vision and Pattern Recognition (CVPR)}.

\bibitem[{Mensink et~al.(2023)Mensink, Uijlings, Castrejon, Goel, Cadar, Zhou, Sha, Araujo, and Ferrari}]{mensink23iccv}
Thomas Mensink, Jasper Uijlings, Lluis Castrejon, Arushi Goel, Felipe Cadar, Howard Zhou, Fei Sha, Andre Araujo, and Vittorio Ferrari. 2023.
\newblock Encyclopedic {VQA}: Visual questions about detailed properties of fine-grained categories.
\newblock In \emph{ICCV}.

\bibitem[{Radford et~al.(2021)Radford, Kim, Hallacy, Ramesh, Goh, Agarwal, Sastry, Askell, Mishkin, Clark et~al.}]{clip}
Alec Radford, Jong~Wook Kim, Chris Hallacy, Aditya Ramesh, Gabriel Goh, Sandhini Agarwal, Girish Sastry, Amanda Askell, Pamela Mishkin, Jack Clark, and 1 others. 2021.
\newblock Learning transferable visual models from natural language supervision.
\newblock In \emph{International conference on machine learning (ICML)}.

\bibitem[{Robertson et~al.(2009)Robertson, Zaragoza et~al.}]{bm25}
Stephen Robertson, Hugo Zaragoza, and 1 others. 2009.
\newblock The probabilistic relevance framework: Bm25 and beyond.
\newblock In \emph{Foundations and Trends{\textregistered} in Information Retrieval}.

\bibitem[{Sim et~al.(2025)Sim, Zhang, Dai, and Fang}]{sim2025can}
Mong~Yuan Sim, Wei~Emma Zhang, Xiang Dai, and Biaoyan Fang. 2025.
\newblock Can vlms actually see and read? a survey on modality collapse in vision-language models.
\newblock In \emph{Findings of the Association for Computational Linguistics: ACL 2025}, pages 24452--24470.

\bibitem[{van~der Maaten and Hinton(2008)}]{JMLR:v9:vandermaaten08a}
Laurens van~der Maaten and Geoffrey Hinton. 2008.
\newblock \href {http://jmlr.org/papers/v9/vandermaaten08a.html} {Visualizing data using t-sne}.
\newblock \emph{Journal of Machine Learning Research}, 9(86):2579--2605.

\bibitem[{Wang et~al.(2020)Wang, Tran, and Feiszli}]{Wang_2020_CVPR}
Weiyao Wang, Du~Tran, and Matt Feiszli. 2020.
\newblock What makes training multi-modal classification networks hard?
\newblock In \emph{Proceedings of the IEEE/CVF Conference on Computer Vision and Pattern Recognition (CVPR)}.

\bibitem[{Wu et~al.(2025)Wu, Tang, Zheng, and Jiang}]{wu2025language}
Huyu Wu, Meng Tang, Xinhan Zheng, and Haiyun Jiang. 2025.
\newblock When language overrules: Revealing text dominance in multimodal large language models.
\newblock \emph{arXiv preprint arXiv:2508.10552}.

\bibitem[{Wu et~al.(2024)Wu, Dadu, Tustison, Avants, Nalls, Sun, and Faghri}]{wu2024multimodal}
Zhenbang Wu, Anant Dadu, Nicholas Tustison, Brian Avants, Mike Nalls, Jimeng Sun, and Faraz Faghri. 2024.
\newblock \href {https://openreview.net/forum?id=Je5SHCKpPa} {Multimodal patient representation learning with missing modalities and labels}.
\newblock In \emph{The Twelfth International Conference on Learning Representations}.

\bibitem[{Xiao et~al.(2023)Xiao, Liu, Zhang, and Muennighoff}]{bge}
Shitao Xiao, Zheng Liu, Peitian Zhang, and Niklas Muennighoff. 2023.
\newblock \href {https://arxiv.org/abs/2309.07597} {C-pack: Packaged resources to advance general chinese embedding}.
\newblock \emph{Preprint}, arXiv:2309.07597.

\bibitem[{Xiong et~al.(2021)Xiong, Xiong, Li, Tang, Liu, Bennett, Ahmed, and Overwijk}]{ANCE}
Lee Xiong, Chenyan Xiong, Ye~Li, Kwok-Fung Tang, Jialin Liu, Paul~N. Bennett, Junaid Ahmed, and Arnold Overwijk. 2021.
\newblock Approximate nearest neighbor negative contrastive learning for dense text retrieval.
\newblock In \emph{International Conference on Learning Representations (ICLR)}.

\bibitem[{Zhai et~al.(2023)Zhai, Mustafa, Kolesnikov, and Beyer}]{zhai2023sigmoid}
Xiaohua Zhai, Basil Mustafa, Alexander Kolesnikov, and Lucas Beyer. 2023.
\newblock Sigmoid loss for language image pre-training.
\newblock In \emph{Proceedings of the IEEE/CVF international conference on computer vision}, pages 11975--11986.

\bibitem[{Zhang et~al.(2025{\natexlab{a}})Zhang, Moiseev, Ainslie, Suganthan, Ma, Bhupatiraju, Lebron, Firat, Joulin, and Dong}]{zhang2025encoderdecodergemmaimprovingqualityefficiency}
Biao Zhang, Fedor Moiseev, Joshua Ainslie, Paul Suganthan, Min Ma, Surya Bhupatiraju, Fede Lebron, Orhan Firat, Armand Joulin, and Zhe Dong. 2025{\natexlab{a}}.
\newblock \href {https://arxiv.org/abs/2504.06225} {Encoder-decoder gemma: Improving the quality-efficiency trade-off via adaptation}.
\newblock \emph{Preprint}, arXiv:2504.06225.

\bibitem[{Zhang et~al.(2022)Zhang, Chu, Ma, Zhu, Wang, Wang, and Zhao}]{10.1145/3534678.3539388}
Chaohe Zhang, Xu~Chu, Liantao Ma, Yinghao Zhu, Yasha Wang, Jiangtao Wang, and Junfeng Zhao. 2022.
\newblock \href {https://doi.org/10.1145/3534678.3539388} {M3care: Learning with missing modalities in multimodal healthcare data}.
\newblock In \emph{Proceedings of the 28th ACM SIGKDD Conference on Knowledge Discovery and Data Mining}, KDD '22, page 2418–2428, New York, NY, USA. Association for Computing Machinery.

\bibitem[{Zhang et~al.(2025{\natexlab{b}})Zhang, Zhang, Xie, Li, Dai, Long, Xie, Zhang, Li, and Zhang}]{zhang2025gmeimprovinguniversalmultimodal}
Xin Zhang, Yanzhao Zhang, Wen Xie, Mingxin Li, Ziqi Dai, Dingkun Long, Pengjun Xie, Meishan Zhang, Wenjie Li, and Min Zhang. 2025{\natexlab{b}}.
\newblock \href {https://arxiv.org/abs/2412.16855} {Gme: Improving universal multimodal retrieval by multimodal llms}.
\newblock \emph{Preprint}, arXiv:2412.16855.

\bibitem[{Zhang et~al.(2025{\natexlab{c}})Zhang, Li, Long, Zhang, Lin, Yang, Xie, Yang, Liu, Lin, Huang, and Zhou}]{zhang2025qwen3embeddingadvancingtext}
Yanzhao Zhang, Mingxin Li, Dingkun Long, Xin Zhang, Huan Lin, Baosong Yang, Pengjun Xie, An~Yang, Dayiheng Liu, Junyang Lin, Fei Huang, and Jingren Zhou. 2025{\natexlab{c}}.
\newblock \href {https://arxiv.org/abs/2506.05176} {Qwen3 embedding: Advancing text embedding and reranking through foundation models}.
\newblock \emph{Preprint}, arXiv:2506.05176.

\bibitem[{Zhang et~al.(2025{\natexlab{d}})Zhang, Zhao, Hu, Jiao, Jiang, Miao, and Nguyen}]{zhang2025learningcompressunlockingpotential}
Yeqin Zhang, Yizheng Zhao, Chen Hu, Binxing Jiao, Daxin Jiang, Ruihang Miao, and Cam-Tu Nguyen. 2025{\natexlab{d}}.
\newblock \href {https://arxiv.org/abs/2511.17129} {Learning to compress: Unlocking the potential of large language models for text representation}.
\newblock \emph{Preprint}, arXiv:2511.17129.

\bibitem[{Zheng et~al.(2025)Zheng, Liao, Fu, Lei, Lyu, Jiang, Ren, Chen, Wang, Li et~al.}]{zheng2025mllms}
Xu~Zheng, Chenfei Liao, Yuqian Fu, Kaiyu Lei, Yuanhuiyi Lyu, Lutao Jiang, Bin Ren, Jialei Chen, Jiawen Wang, Chengxin Li, and 1 others. 2025.
\newblock Mllms are deeply affected by modality bias.
\newblock \emph{arXiv preprint arXiv:2505.18657}.

\bibitem[{Zhou et~al.(2024{\natexlab{a}})Zhou, Liu, Xiao, Zhao, and Xiong}]{zhou-etal-2024-vista}
Junjie Zhou, Zheng Liu, Shitao Xiao, Bo~Zhao, and Yongping Xiong. 2024{\natexlab{a}}.
\newblock \href {https://doi.org/10.18653/v1/2024.acl-long.175} {{VISTA}: Visualized text embedding for universal multi-modal retrieval}.
\newblock In \emph{Proceedings of the 62nd Annual Meeting of the Association for Computational Linguistics (Volume 1: Long Papers)}, pages 3185--3200, Bangkok, Thailand. Association for Computational Linguistics.

\bibitem[{Zhou et~al.(2024{\natexlab{b}})Zhou, Mei, Li, Liu, Xiong, Liu, Gu, and Yu}]{zhou-etal-2024-marvel}
Tianshuo Zhou, Sen Mei, Xinze Li, Zhenghao Liu, Chenyan Xiong, Zhiyuan Liu, Yu~Gu, and Ge~Yu. 2024{\natexlab{b}}.
\newblock \href {https://doi.org/10.18653/v1/2024.acl-long.783} {{MARVEL}: Unlocking the multi-modal capability of dense retrieval via visual module plugin}.
\newblock In \emph{Proceedings of the 62nd Annual Meeting of the Association for Computational Linguistics (Volume 1: Long Papers)}, pages 14608--14624, Bangkok, Thailand. Association for Computational Linguistics.

\end{thebibliography}

\clearpage

\appendix

\section{Appendix}



\subsection{Detail Introduction of Datasets}
\label{sec:appendix-data}

\paragraph{WebQA} This dataset contains images and passage snippets that are crawled from the general Web and Wikipedia. It is a multi-modal retrieval that includes both text-to-text (T2T) retrieval, where a text query retrieves text documents, and text-to-image (T2I) retrieval, where a text query retrieves image documents. The ratio of these two types of queries in the WebQA dataset is 1:1. Its corpus is a hybrid corpus containing both images and text, with image documents all having captions. We use the same WebQA dataset as \UniVLDR{} and \Marvel{} do.

\paragraph{EVQA} Encyclopedic-VQA is a visual question answering (VQA) dataset that requires the ability to understand and reason about detailed encyclopedic knowledge. The dataset is larger in scale, and the questions are designed to be truly multimodal. This means that answers cannot be based solely on images or text alone. Each answer is annotated with supporting evidence from the corresponding Wikipedia section. 

We use the processed EVQA dataset from \textit{BByrneLab/M2KR} and create an EVQA+ dataset based on EVQA, merging WebQA's T2T task into it and making it an any-modality-to-text-corpus version. The captions of the query-side images in the EVQA+ dataset are taken from the title of its ground truth wiki documentation.

\subsection{Implementation Details of Baseline}
\label{sec:appendix1}

\paragraph{\UniVLDR{} Training} 
For training CLIP-DPR—the first stage checkpoint \UniVLDR{}—we start from the ViT-B/32 version of CLIP and continuously train CLIP on the WebQA+ dataset or EVQA+ dataset with in-batch negatives. Here caption ratio of datasets in training is set to 1, which is equivalent to the original WebQA dataset, to align with the original result.
We truncate texts with the max length of 77 and set accumulate step as 1, batch size to 64, max training epoch to 20, and the temperature hyperparameter $\tau$ = 0.01. The learning rate is $5\times10^{-6}$ for the WebQA+ dataset and $1\times10^{-5}$ for the EVQA+ dataset. The cosine annealing strategy is used to schedule the learning rate during training. 

In the second stage, for training \UniVLDR{}, we retrieve Top 100 documents using CLIP-DPR and sample two hard negatives of different modalities (k = 1) from these candidates. All models are tuned with AdamW optimizer, are evaluated per 500 steps, and set early stop step as 5. Training is conducted on a single A100.

\paragraph{\Marvel{} Training} \Marvel{} model is based on a pre-trained T5 model and uses CLIP's visual encoder as an image tokenizer, and 128 max text token length and 49 image token length are set. Following the original setting in paper, we trained on a single A100, using a batch size of 64 and a learning rate of $5\times10^{-6}$ for WebQA+ datasets and a batch size of 64 and a learning rate of $1\times10^{-5}$ for EVQA+ datasets. Caption ratio of datasets in training is set to 1 to align with the original result. The temperature hyperparameter $\tau$ = 0.01. Early stopping is implemented like \UniVLDR{}.

We follow the two-stage training of \Marvel{}: In the first stage, we train the model using only in-batch negative examples. In the second stage, we use both self-mined hard negatives and in-batch negatives to obtain the \Marvel{}-ANCE model, similar to that of \UniVLDR{}.

\paragraph{VISTA Training} VISTA model is based on a pre-trained BGE-base model and uses EVA-CLIP's visual encoder as an image tokenizer, and 512 max text token length and 196 image token length are set. For a fair comparison, we replaced EVA-CLIP with CLIP-B-32, and named VISTA* in paper. The performance differences between VISTA* and the original VISTA on the WebQA+ dataset are shown in Table \ref{tab:vista}. Our proposed method (MiMIC-ANCE) achieves superior performance even with smaller-parameter ViT models. This indicates that VISTA struggles with T2I retrieval, especially, when the image encoder is replaced by CLIP-B-32.

\begin{table}[!]
    \centering
    \resizebox{0.48\textwidth}{!}{%
        \begin{tabular}{l|ccc} 
        \toprule
        Method & T2T & T2I & T2All \\
        \midrule
        \textbf{VISTA*(CLIP-B-32)} & 82.9 &  5.1 & 43.5\\
        \textbf{VISTA(EVA-CLIP-16)} & 82.5 & 42.8 & 62.9 \\
        \textbf{MiMIC-ANCE(CLIP-B-32)} & 80.2 & 59.4 & 69.6 \\
            \bottomrule
        \end{tabular}%
    }
    \caption{R@20 performance of VISTA, VISTA*, and \ourmethod{}-ANCE on different modality retrieval tasks of WebQA+ Dataset.}
    \label{tab:vista}
\end{table}

Following the original setting in paper, we trained on a single A100, using a batch size of 128 and a learning rate of $1\times10^{-5}$ for WebQA+ datasets and a batch size of 64 and a learning rate of $1\times10^{-5}$ for EVQA+ datasets. Caption ratio of datasets in training is set to 1.

\subsection{Implementation Details of MiMIC}
\label{sec:appendix2}

During training MiMIC, we employ the T5 text encoder and CLIP image encoder initialized with the t5-ance checkpoint from OpenMatch and clip-vit-base-patch32 checkpoint from OpenAI, truncate the text with the max length of 128 and set the batch size to 64, learning rate=$5e-6$, max training epoch to 20, and the temperature hyperparameter $\tau$ = 0.01. And cosine annealing strategy is used to schedule the learning rate during training. All models are tuned with AdamW optimizer, are evaluated per 500 steps, and set early stop step as 5. Like the Implementation of \UniVLDR{} and \Marvel{}. The parameter distribution of each component in the model is shown in the table \ref{tab:model-para}.

\begin{table}[!]
    \centering
    \resizebox{0.48\textwidth}{!}{%
        \begin{tabular}{c|cccc} 
        \toprule
        & \makecell{Image\\Encoder} & \makecell{Text\\Encoder} & \makecell{Fusion\\Decoder} & Total \\
        \midrule
        Paras & 87M & 109M & 138M & 335M \\
            \bottomrule
        \end{tabular}%
    }
    \caption{The parameter size of MiMIC Model.}
    \label{tab:model-para}
\end{table}

We set the caption ratio to 0.5 of WebQA+ and EVQA+ datasets, meaning that only half of the multi-modal candidates or queries in the batch have captions. Additionally, when calculating the training features for the Multimodal Candidates ${I_d^{VC}}$ from the WebQA+ dataset, we set $\bar{\alpha}$ to 0.1, and when calculating the training features for the Multimodal Queries ${TI_q^{VC}}$ from the EVQA+ dataset, we set $\bar{\alpha}$ to 0.5. 

In our second training stage, we retrieve Top 100 documents using MiMIC and sample one image hard negative (can be have or not have caption) and one text hard negative (k = 1) from these candidates for WebQA+ dataset or sample one text hard negative (k = 1) for EVQA+ dataset. After training using in-batch hard negatives and hard negatives, we get the MiMIC-ANCE model.

\newpage

\subsection{Different Multi-modal Retrieval of EVQA+}
\label{sec:dmr-evqa+}
\begin{figure}[t]
  \centering
  \includegraphics[width=\linewidth]{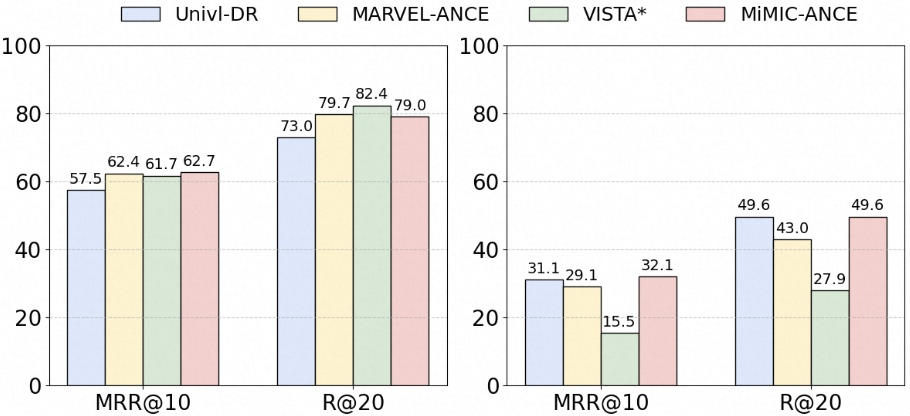}
  \caption{The retrieval performance of T2T (left) and TI2T (right) task separately on EVQA+ dataset using different methods.}
  \label{fig:evqa_single}
\end{figure}

Here we show the performance of different modality subtask on EVQA+ Dataset in Figure \ref{fig:evqa_single}. \Marvel{}-ANCE performs consistently and well in the T2T task, but lags behind MiMIC-ANCE in the TI2T task. While Univl-DR has relatively low metrics in the T2T task, it shows good competitiveness in the TI2T task. Similarly, VISTA*, as on the WebQA+ dataset, exhibits significant differences across different tasks. In the TI2T task, MiMIC-ANCE performs best, ranking first in both MRR@10 (32.1) and R@20 (49.6). This indicates that the model has stronger feature modeling capabilities and retrieval accuracy when handling complex visual-text joint queries.

\end{document}